\documentclass[]{beingbeyond}
\usepackage{enumitem}
\usepackage[toc,page,header]{appendix}

\usepackage{amsmath,amssymb,amsthm,bm}
\usepackage{colortbl}
\usepackage{algorithm}
\usepackage{algpseudocode}
\usepackage{makecell}
\usepackage{adjustbox}
\usepackage{tabularx}
\usepackage{xspace}

\definecolor{BlockA}{RGB}{204,227,255}

\newcommand{\R}{\mathbb{R}}
\newcommand{\E}{\mathbb{E}}
\newcommand{\sg}{\mathrm{sg}}
\newcommand{\cR}{\mathcal{R}}
\newcommand{\cL}{\mathcal{L}}

\newcommand{\DiG}{\texttt{DiG}\xspace}
\newcommand{\DiGRefine}{\texttt{DiG-Refine}\xspace}

\newcommand{\eg}{\textit{e.g.}}

\usepackage{coloredtheorem}
\cthnewtheorem{proposition}{Proposition}
\cthnewtheorem{theorem}{Theorem}

\theoremstyle{definition}

\theoremstyle{remark}

\crefname{equation}{Eq.}{Eqs.}
\Crefname{equation}{Eq.}{Eqs.}
\crefname{theorem}{Theorem}{Theorems}
\Crefname{theorem}{Theorem}{Theorems}
\crefname{proposition}{Proposition}{Propositions}
\Crefname{proposition}{Proposition}{Propositions}
\crefname{section}{Section}{Sections}
\Crefname{section}{Section}{Sections}
\crefname{subsection}{Section}{Sections}
\Crefname{subsection}{Section}{Sections}
\crefname{algorithm}{Algorithm}{Algorithms}
\Crefname{algorithm}{Algorithm}{Algorithms}

\title{Transport Discrepancy as a Reliability Signal for Vision-Language-Action Models}

\author{{\bfseries
Wanpeng Zhang$^{1,3}$~
Ye Wang$^{2,3}$~
Hao Luo$^{1,3}$~
Haoqi Yuan$^{1,3}$~
Yicheng Feng$^{1,3}$ \\
Chaoyi Xu$^{1,3}$~
Sipeng Zheng$^{3}$~
Qin Jin$^{2}$~
Zongqing Lu$^{1,3,\dagger}$
}}

\affiliation{
$^{1}$Peking University \quad
$^{2}$Renmin University of China \quad
$^{3}$BeingBeyond \quad
}

\webpage{\url{https://research.beingbeyond.com/dig}}

\abstract{
Vision-language-action (VLA) models that generate continuous action chunks via flow matching lack an internal signal for judging whether a given prediction is reliable.
Distribution shift and long-horizon rollouts can push backbone representations away from the region the action head decodes reliably, yet the policy has no mechanism to detect or react to this drift.
We observe that the cost of transporting observation features to the action representation in a shared feature space rises precisely when such drift occurs, providing a per-step reliability estimate without extra supervision.
Building on this observation, we propose \DiG (Discrepancy Gate), a lightweight plug-in module for flow-matching VLA policies.
\DiG computes a sliced Wasserstein transport cost between backbone features and the action expert's own input projection, maps it through an exponential gate, and uses the gate to modulate both a residual feature refinement and the training loss.
At inference time, the gate enables \DiGRefine, an iterative refinement process that corrects action chunks before execution.
Experiments on both simulation and real-world scenarios show that \DiG consistently improves success rates, with the largest gains under distribution shift and on long-horizon tasks.
}

\checkdata[Date]{Dec 01, 2025}

\begin{document}

\maketitle

\begingroup
\renewcommand\thefootnote{\fnsymbol{footnote}}
\setcounter{footnote}{0}
\footnotetext[1]{This paper is accepted by ECCV 2026.}
\footnotetext[2]{Correspondence to Zongqing Lu $<$\href{mailto:lu@beingbeyond.com}{lu@beingbeyond.com}$>$.}
\endgroup

\section{Introduction}

Vision-language-action (VLA) models map images and language instructions to robot controls by combining a pretrained vision-language backbone with an action head \cite{brohan2023rt1,kim2024openvla,black2024pi0}.
Recent systems often generate continuous action chunks with diffusion or flow matching inside a dedicated action expert \cite{black2024pi0,intelligence2025pi05,bjorck2025gr00t}.
In flow matching, the action expert learns a time-conditioned velocity field whose ODE \emph{transports} a simple noise distribution into the action distribution.
This transport view is the mechanism behind action generation, and it also provides a natural way to detect when the policy is likely operating out of distribution: the cost of transporting observation features to the action representation rises when they become misaligned.

For a fixed backbone and action head, only part of the backbone representation space can be reliably decoded into correct actions.
This subset behaves like a low-dimensional \emph{decodable manifold}.
Distribution shift and long-horizon rollouts can push the backbone features away from that manifold, so the action head still produces a chunk, but the prediction is poorly conditioned. Errors then compound across steps.
A practical implication is the need for a per-step reliability estimate.
Such a signal can serve two roles.
During training, it can reduce the influence of demonstration pairs that are explained by shortcuts or spurious cues \cite{geirhos2020shortcut,xing2025shortcut}.
At inference time, it can indicate when a predicted chunk is inconsistent with the current observation and should be corrected before execution.

This paper introduces \DiG (Discrepancy Gate), a plug-in module for flow-matching VLA policies.
\DiG turns the transport view of flow matching into an \emph{internal reliability signal}.
It computes a transport discrepancy between (i) the backbone feature distribution of the current observation and (ii) a compact action representation obtained by the same action expert.
The discrepancy is instantiated as a sliced 2-Wasserstein transport cost \cite{bonneel2015sliced,kolouri2019generalized}.
A monotone mapping produces a gate value that modulates a lightweight residual refinement of the backbone features.
The same gate also reweights the flow-matching loss.
During backpropagation, $\sg(\cdot)$ is applied only to the gate.
This blocks direct optimization of the discrepancy while keeping learning signals from the original flow-matching loss.
At deployment, \DiGRefine runs a short inference-time refinement loop that reuses the same gate with predicted chunks.
The module attaches to the context features consumed by the action expert.
It is compatible with both shared-transformer designs that use prefix KV caches and two-stage designs with a separate action expert.

The contributions of our paper are as follows: 1) A transport-based reliability signal for flow-matching VLA policies, realized as a lightweight module and a matching inference-time procedure that improve robustness without changing the flow-matching target. 2) A theoretical analysis that connects the exponential gate to a local manifold noise model (Proposition~\ref{prop:posterior}) and formalizes how gating suppresses shortcut components under a mixture contamination model (Theorem~\ref{thm:mass}). 3) Extensive experiments on both simulation and real-robot tasks, including perturbation and sensitivity analyses, to verify the effectiveness of \DiG.

\section{Related Work}

VLA policies build on progress in vision-language models (VLMs).
Large multimodal backbones couple language reasoning \cite{touvron2023llama,touvron2023llama2,bai2023qwen} with strong visual encoders \cite{radford2021learning,zhai2023sigmoid,feng2024videoorion,luo2025openmmego} and exhibit broad instruction-following behavior \cite{alayrac2022flamingo,liu2024improved,liu2023visual,lu2023empirical,zhang2025beingvl0,zhang2025beingvl05,zhu2023minigpt,wang2024qwen2}.
VLA work transfers these representations to robot control.
RT-1 and RT-2 scale transformer policies with web data and large robot datasets \cite{brohan2023rt1,brohan2023rt2}.
OpenVLA provides an open 7B-scale model and benchmark suite \cite{kim2024openvla,kim2025fine}.
Other lines explore modular and multi-embodiment policies \cite{octomodel2024,driess2023palm}.
Several recent systems generate actions with diffusion or flow matching to model multimodality \cite{black2024pi0,intelligence2025pi05,bjorck2025gr00t,chi2023dp}.
GR-2 and GR-3 leverage video pretraining for robot control \cite{cheang2024gr2vla,cheang2025gr}.
SpatialVLA and VIPA-VLA incorporate spatial representations \cite{qu2024spatialvla,feng2026spatialaware}.
CoT-VLA adds visual chain-of-thought reasoning \cite{zhao2025cotvla}.
Otter focuses on fine-grained visual features \cite{huang2025otter}.
OneTwoVLA studies adaptive reasoning in a unified architecture \cite{lin2025onetwovla}.
Being-H0 incorporates physical instruction tuning for better motion understanding \cite{beingbeyond2025beingh0}, and Being-H0.5 scales human-centric learning with a unified action space for cross-embodiment robot control \cite{beingbeyond2026beingh05}. Several recent systems use predictive representations to support future-aware robot control \cite{luo2026jointalignedlatentaction,beingbeyond2026beingh07,zhang2026conservative,zhang2025dreamvla,sun2026vla,zhang2026disentangled}.
These systems still struggle under distribution shift, especially in long-horizon tasks, and typically lack an internal reliability estimate \cite{wang2026rethinking}.

Flow matching provides the action-generation mechanism for many VLA policies. It learns a time-conditioned velocity field without simulating a forward process \cite{lipman2023flow}.
Related work includes conditional flow matching \cite{tong2023conditional}, rectified flow \cite{liu2022flow}, improved probability paths \cite{liu2024improved}, and stochastic interpolants \cite{albergo2023stochastic}.
Flow matching connects to continuous normalizing flows \cite{chen2018neural,grathwohl2018ffjord,papamakarios2021normalizing}.
Optimal transport (OT) offers a metric view of distribution comparison \cite{villani2009optimal,peyre2019computational}.
Computational OT methods include Sinkhorn divergence \cite{cuturi2013sinkhorn} and sliced Wasserstein distances \cite{bonneel2015sliced,kolouri2019generalized}.
OT-CFM uses OT to improve flow matching trajectories \cite{pooladian2023multisample,tong2024improving}.
Our focus differs: transport costs serve not as a way to alter the probability path but as an auxiliary reliability signal that modulates backbone features.

Robustness in robot learning is a long-standing theme.
Domain randomization broadens training support \cite{tobin2017domain}.
Domain adaptation aligns source and target distributions \cite{tzeng2020adapting}.
Invariance and out-of-distribution generalization are studied in broader machine learning \cite{krueger2021out}.
Shortcut learning is a recurring failure mode \cite{geirhos2020shortcut}.
Non-stationary and spurious correlations have been analyzed in robotics benchmarks \cite{zhang2024tackling,xing2025shortcut,feng2022factored}.
This work complements these directions with a mechanism that reacts to drift at inference time through an internal transport-based signal.

\section{Preliminaries}
\label{sec:prelim}

A VLA policy receives observations $o$, language instructions, and proprioceptive states.
A vision-language backbone processes $o$ and produces context features $H = (h_1,\ldots,h_T) \in \R^{T\times d}$, where $T$ is the sequence length and $d$ is the feature dimension.
$H$ is the interface consumed by the action expert; depending on architecture, it can be the transformer hidden states or KV-cached activations.

The action head predicts an action chunk $a = (a_1,\ldots,a_K)\in \R^{K\times d_a}$ with horizon $K$ and action dimension $d_a$.
Flow matching generates $a$ by learning a conditional velocity field $v_\theta(x,t\mid H)$ over $x\in\R^{K\times d_a}$ and $t\in[0,1]$.
The training interpolant is $x_t = (1-t)\,\varepsilon + t\,a$ with noise $\varepsilon\sim\mathcal{N}(0,I)$ and $t\sim\mathcal{U}[0,1]$.
The flow-matching objective regresses $v_\theta$ to the conditional target velocity~\cite{lipman2023flow}:
\begin{equation}
\label{eq:fm}
\cL_{\mathrm{FM}}(\theta)
\;=\;
\E_{t,\varepsilon,a}\Big[\big\|v_\theta(x_t,t\mid H) - u(x_t,t,a)\big\|^2\Big].
\end{equation}

Optimal transport defines a cost between two measures $\mu,\nu$ over $\R^d$.
The squared 2-Wasserstein distance is
\begin{equation}
W_2^2(\mu,\nu)
\;=\;
\inf_{\gamma\in\Gamma(\mu,\nu)}\;
\int \|x-y\|^2\,d\gamma(x,y),
\end{equation}
where $\Gamma(\mu,\nu)$ is the set of couplings with marginals $\mu$ and $\nu$ \cite{villani2009optimal}.
Sliced Wasserstein distances approximate $W_2$ by averaging 1D Wasserstein costs along $M$ random projections on the unit sphere \cite{bonneel2015sliced,kolouri2019generalized}; $M$ controls the trade-off between variance and computation.

\begin{figure}[t]
\centering
\includegraphics[width=\linewidth]{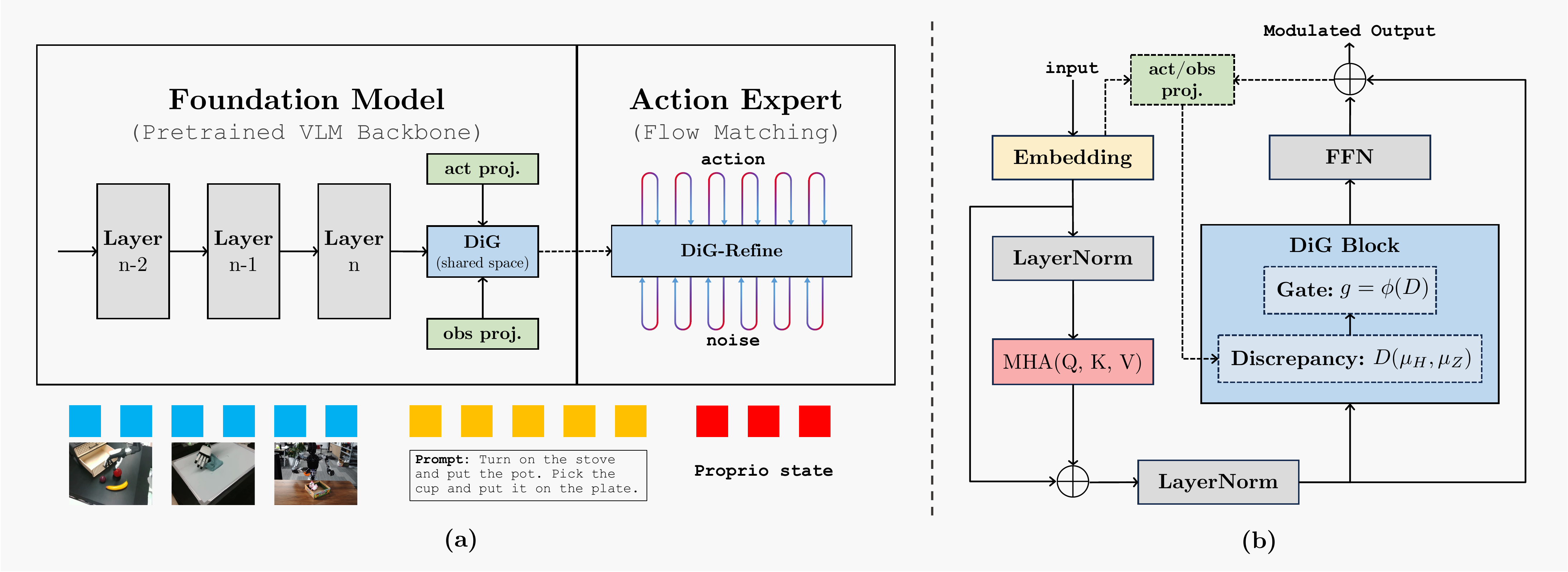}
\caption{\textbf{Overview of \DiG.}
The backbone produces observation features $H$.
The action projection $f$ (the input projector from the action expert) maps the action chunk into the same feature space, yielding $Z$.
A sliced Wasserstein cost computes the transport discrepancy $D$ between $H$ and $Z$.
An exponential mapping converts $D$ into a gate $g$, which modulates a residual refinement to produce $\tilde{H}$.
$\tilde{H}$ is then passed to the action expert for action generation.
The same gate also reweights the training loss.}
\label{fig:framework}
\end{figure}

\section{Discrepancy gating from transport reliability}
\label{sec:method}

This section introduces \DiG as a transport-based reliability module.
The design follows one principle: estimate whether the current observation representation lies in the action expert's decodable region, and use this estimate to gate both training and inference.

\subsection{Transport discrepancy as an internal reliability signal}
\label{sec:reliability-signal}

The action expert's input projection $f$ maps action chunks into the same feature space as the backbone output $H$.
A transport cost between these two representations quantifies their compatibility: low cost indicates that $H$ lies in a region the action expert can decode reliably, while high cost signals drift away from that region.
This cost serves as a per-step reliability estimate without extra supervision.
During training, it reduces the influence of shortcut-like pairs.
At test time, it flags when a predicted chunk is inconsistent with the current observation.

\subsection{\DiG module and training}
\label{sec:dig-train}

Figure~\ref{fig:framework} gives an overview of \DiG.
The module attaches to the backbone context features consumed by the action expert.
In a shared-transformer design, it refines the hidden states before the action expert.
In a two-stage design, it refines the conditioning representation passed to a separate diffusion transformer.

\noindent\textbf{Action representation in the discrepancy space.}
The transport discrepancy compares observation representations against an action representation expressed in the same feature coordinates.
The action projection $f:\R^{d_a}\rightarrow\R^{d}$ is the input projection layer already present inside the action expert; it maps (noised) actions into the expert's internal feature dimension.
Because $f$ sits on the main action-generation path, it receives gradients from the standard flow-matching loss without any additional training signal.
\DiG reuses $f$ directly: each action step yields $z_k = f(a_k)\in\R^d$, and a mean-pooled centroid $\bar z = \tfrac{1}{K}\textstyle\sum_{k=1}^K z_k$ summarizes the chunk.
The centroid is broadcast to length $T$ to form $Z=(\bar z,\ldots,\bar z)\in\R^{T\times d}$, removing action ordering effects and focusing the discrepancy on overall feature alignment.
The centroid is used only inside the discrepancy branch; the flow-matching head still receives and models the full action chunk.
This design keeps the reliability signal focused on chunk-level compatibility.

\noindent\textbf{Transport discrepancy.}
Empirical measures $\mu_H = \frac{1}{T}\sum_{i=1}^T \delta_{h_i}$ and $\mu_Z = \frac{1}{T}\sum_{i=1}^T \delta_{\bar z}$ are constructed from the feature sequences.
The transport discrepancy is a sliced 2-Wasserstein cost
\begin{equation}
D(\mu_H,\mu_Z)
\;=\;
\frac{1}{M}\sum_{m=1}^M W_2^2\big(\langle \theta_m,\mu_H\rangle,\langle \theta_m,\mu_Z\rangle\big),
\label{eq:sw}
\end{equation}
Because $\mu_Z$ is a point mass, each projected term admits a closed-form expression.
For a unit direction $\theta$, define $u_i=\theta^\top h_i$ and $v=\theta^\top\bar z$.
Then
\begin{equation}
D_\theta
\;=\;
W_2^2\big(\langle \theta,\mu_H\rangle,\langle \theta,\mu_Z\rangle\big)
\;=\;
\frac{1}{T}\sum_{i=1}^T (u_i-v)^2,
\label{eq:1d-closed}
\end{equation}
where $\langle \theta,\mu\rangle$ denotes the pushforward of $\mu$ under $x\mapsto\theta^\top x$.
The point-mass form is the key estimator choice: it makes the discrepancy cheap, but it also keeps the reliability branch focused on compatibility between the current observation representation and the action expert's own coordinates.

\begin{figure}[tb]
  \centering
  \includegraphics[width=.5\linewidth]{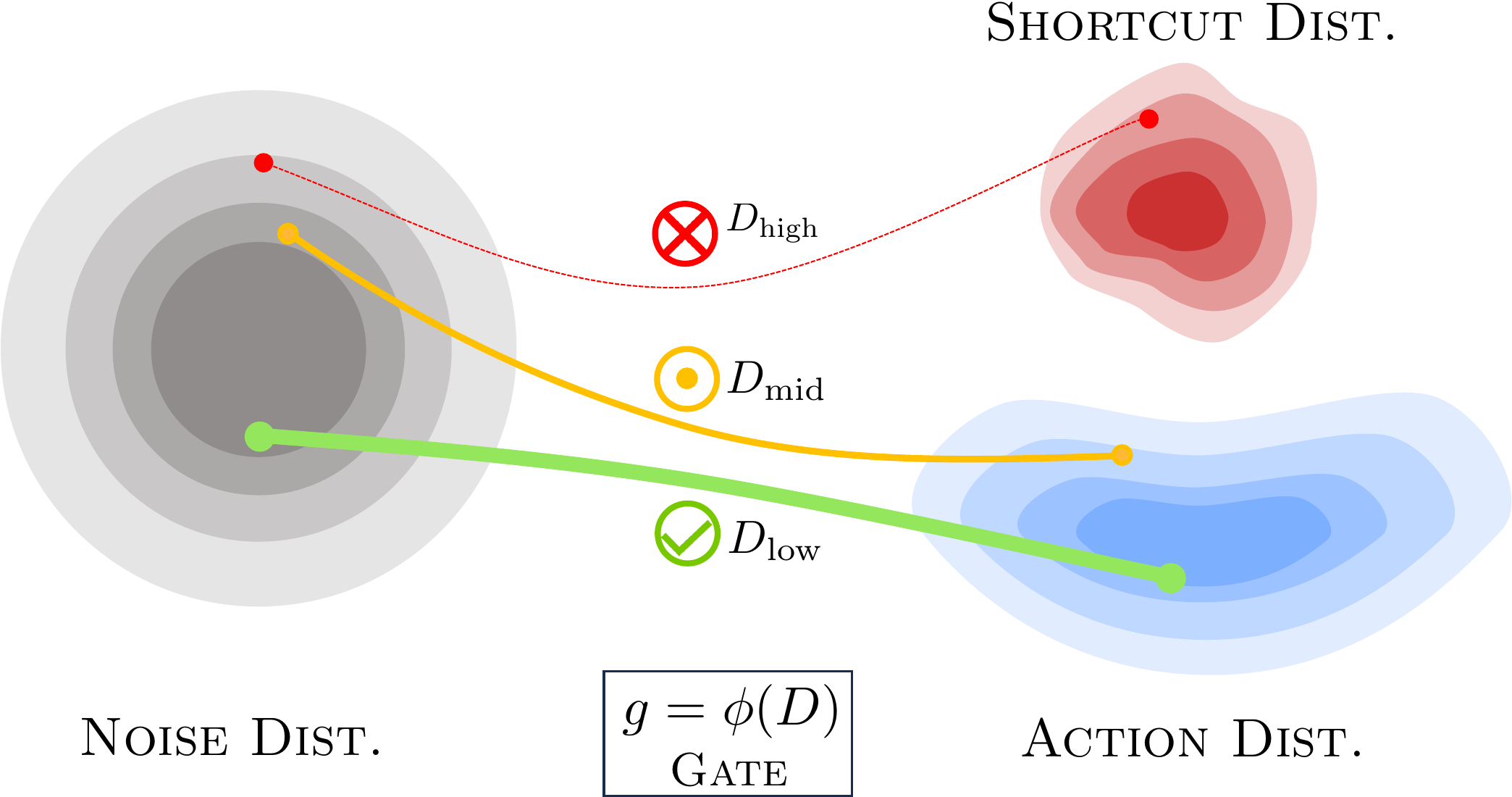}
  \caption{\textbf{Transport discrepancy and compatibility.}
  The curved surface denotes the decodable region of representation space.
  Green points denote coherent observation-action pairs; red points denote shifted or shortcut-correlated pairs.}
  \label{fig:digflow-intuition}
\end{figure}

Figure~\ref{fig:digflow-intuition} illustrates this compatibility view.
Coherent observation-action pairs lie near the decodable region and require little transport; shifted or shortcut-correlated pairs lie farther away and should receive less trust.

A local noise model makes the desired gate shape explicit.
Along one random projection direction $\theta$, the discrepancy term $D_\theta$ in Eq.~(\ref{eq:1d-closed}) is the average squared projected residual between observation tokens and the action centroid.
If coherent pairs concentrate near the decodable region, then residuals in local normal directions can be approximated by small Gaussian perturbations.
Under this model, large residual energy has an exponential tail.

\begin{cthproposition}{Exponential gate as a posterior proxy}
\label{prop:posterior}
Consider one projected term $D_\theta$ in Eq.~(\ref{eq:1d-closed}).
Assume the projected residuals $u_i-v$ are i.i.d.\ $\mathcal{N}(0,\sigma^2)$.
Then $D_\theta$ is a scaled chi-square random variable and its density satisfies
\begin{equation}
p(D_\theta\mid\sigma^2)
\;=\;
C\,D_\theta^{T/2-1}\exp\!\left(-\frac{T}{2\sigma^2}D_\theta\right),
\qquad
D_\theta>0,
\label{eq:Dtheta-gamma}
\end{equation}
where $C$ is a constant independent of $D_\theta$.
\end{cthproposition}

Proposition~\ref{prop:posterior} gives the gate a local statistical interpretation.
High transport discrepancy is exponentially less compatible with the coherent-pair model, so $g=\exp(-\tau D)$ preserves the desired monotone reliability ordering while $\tau$ controls calibration.
The density also contains a polynomial prefactor, but this term does not change the ordering and is dominated in the large-discrepancy tail where the gate should be small.
Aggregating many slices preserves the same form up to temperature scaling, since $D=\tfrac{1}{M}\sum_m D_{\theta_m}$ combines per-slice exponential factors into $\exp(-\tau D)$.

This argument is local rather than a global guarantee about all feature sequences.
If token residuals are correlated, $D_\theta$ becomes a quadratic form instead of a scaled chi-square variable, but its expectation still tracks the residual scatter,
\begin{equation}
\E_\theta[D_\theta]
= \frac{1}{d}\cdot\frac{1}{T}\sum_{i=1}^T \|h_i-\bar z\|^2
= \frac{1}{d}\,\mathrm{tr}(\hat\Sigma_{\mathrm{res}}).
\end{equation}
Correlations mainly affect calibration of $\tau$, while the monotone relationship between feature dispersion and gate value remains the intended signal.
Averaging over $M$ independent slices reduces estimator variance as $\mathrm{Var}(D)=\mathrm{Var}(D_\theta)/M$, which is why moderate projection budgets are sufficient in Figure~\ref{fig:hyper-sens}.

\subsection{Efficient action-interface estimator}\label{sec:estimator-projection}

The point-mass action target above is also what makes the estimator lightweight.
Because $\mu_Z$ is a point mass, Eq.~(\ref{eq:1d-closed}) requires only mean squared projected deviations.
The sliced estimator therefore costs $O(TM)$ for $T$ context tokens and $M$ projection directions, with no sorting or Sinkhorn iterations.
The projection directions are resampled during the forward pass and are not learned.
This keeps \DiG attached to the backbone-action interface rather than introducing a separate transport network.

The centroid should be read as a compact diagnostic target, not as a replacement for the action model.
It asks whether the observation tokens $H$ lie in the region that the action expert can decode from the current action-side representation; the full temporal and multimodal structure of the chunk remains modeled by the flow-matching action head.

During flow-matching training, the action expert consumes the noised interpolant $x_t$ on its generation path, while the discrepancy branch forms $Z$ from the clean demonstrated chunk.
This separates reliability estimation from the auxiliary noise variable $t$ and ties the gate to chunk semantics.
Since the same action-side projector $f$ is reused, the discrepancy branch does not add a second action embedding space or an external representation target.

\noindent\textbf{Gate and residual path.}
The local compatibility model above motivates an exponential gate.
The discrepancy is mapped to a gate value
\begin{equation}
g \;=\; \phi(D) \;=\; \max\{g_{\min},\exp(-\tau D)\},
\label{eq:gate}
\end{equation}
where $g_{\min}\in(0,1)$ prevents complete suppression and $\tau$ controls sensitivity.
For training, we write $\bar g=\sg(g)$ when the gate enters the action-generation path; at inference $\bar g=g$.
The forward value is unchanged by this notation.
Only the backward path from $D$ to $g$ is blocked.

A lightweight residual operator $\cR$ then refines the context features:
\begin{equation}
\tilde H
\;=\;
H + \lambda\, \bar g\, \cR(H),
\label{eq:residual}
\end{equation}
with strength $\lambda>0$.
$\cR$ is a single linear map applied to the normalized context features consumed by the action expert, so the module remains independent of a particular normalization design.

\noindent\textbf{Training objective and gradient routing.}
Training uses the ground-truth action chunk $a^{\mathrm{gt}}$ to compute $Z$ and $g$.
The flow-matching loss is reweighted as
\begin{equation}
J(\theta)
\;=\;
\E\Big[\bar g\,\cL_{\mathrm{FM}}(\theta; \tilde H, a^{\mathrm{gt}})\Big].
\label{eq:train-obj}
\end{equation}
The target velocity $u(x_t,t,a)$ is unchanged; \DiG changes the sample weight and the conditioned representation, not the flow-matching target.
Gradients still pass through $\cL_{\mathrm{FM}}$, $H$, $\cR(H)$, and the action expert, including the projector $f$.
What is blocked is only the direct path $D\rightarrow g$.
This prevents the policy from making the diagnostic score artificially small while preserving the normal action-learning signal.

The loss weight also has a distributional effect.
Training data can be viewed as a mixture of coherent pairs and shortcut-correlated or shifted pairs.
Let $P_{\mathrm{coh}}$ denote the coherent distribution and $P_{\mathrm{sh}}$ the shortcut distribution, with
\begin{equation}
P \;=\; (1-\rho)\,P_{\mathrm{coh}} + \rho\,P_{\mathrm{sh}},
\qquad
\rho\in(0,1).
\label{eq:mixture}
\end{equation}
Applying the gate to the loss is equivalent to training under the reweighted distribution
\begin{equation}
Q(dH\,da) \;\propto\; g(H,a)\,P(dH\,da).
\label{eq:Qdef}
\end{equation}
If coherent samples receive larger gates than shortcut samples, the effective training distribution moves closer to the coherent component.

\begin{cththeorem}{Gating increases coherent mass}
\label{thm:mass}
Assume the mixture model in Eq.~(\ref{eq:mixture}).
Consider an idealized \emph{component-wise} gate that is constant on each mixture component.
It takes value $g_1$ on samples drawn from $P_{\mathrm{coh}}$ and value $g_0$ on samples drawn from $P_{\mathrm{sh}}$, with $g_1>g_0>0$.
Define $Q$ via Eq.~(\ref{eq:Qdef}).
Then $Q$ is a mixture of the same components,
\begin{equation}
Q \;=\; (1-\rho_Q)\,P_{\mathrm{coh}} + \rho_Q\,P_{\mathrm{sh}},
\end{equation}
with shortcut weight
\begin{equation}
\rho_Q
\;=\;
\frac{\rho\,g_0}{(1-\rho)\,g_1 + \rho\,g_0}
\;<\;
\rho.
\label{eq:rhoQ}
\end{equation}
Moreover, for any loss $\ell\in[0,1]$ and any parameter $\theta$,
\begin{equation}
\big|\,\E_{(H,a)\sim Q}[\ell(\theta)] - \E_{(H,a)\sim P_{\mathrm{coh}}}[\ell(\theta)]\,\big|
\;\le\;
\rho_Q.
\label{eq:coh-bound}
\end{equation}
\end{cththeorem}

Theorem~\ref{thm:mass} says that \DiG does not need to identify every shortcut perfectly.
It only needs to assign coherent pairs larger expected weight than shortcut pairs.
Then the effective training distribution contains less shortcut mass than the original data distribution, and the worst-case gap to the coherent distribution is controlled by the remaining shortcut fraction $\rho_Q$.
In the actual model, $g(H,a)$ is continuous rather than component-wise; the same mass-shift logic applies whenever
\begin{equation}
\E_{P_{\mathrm{coh}}}[g] \;>\; \E_{P_{\mathrm{sh}}}[g],
\qquad
\rho_Q
=
\frac{\rho\,\E_{P_{\mathrm{sh}}}[g]}
{(1-\rho)\,\E_{P_{\mathrm{coh}}}[g]+\rho\,\E_{P_{\mathrm{sh}}}[g]}.
\end{equation}
Figure~\ref{fig:transport-curve} shows that the discrepancy remains non-collapsed during training, so the continuous gate retains useful separation in practice.
Proofs of Proposition~\ref{prop:posterior} and Theorem~\ref{thm:mass} are provided in Appendix~\ref{app:main-proofs}.

Figure~\ref{fig:train-infer} summarizes the training and inference pathways.

\begin{figure}[tb]
  \centering
  \includegraphics[width=.64\linewidth]{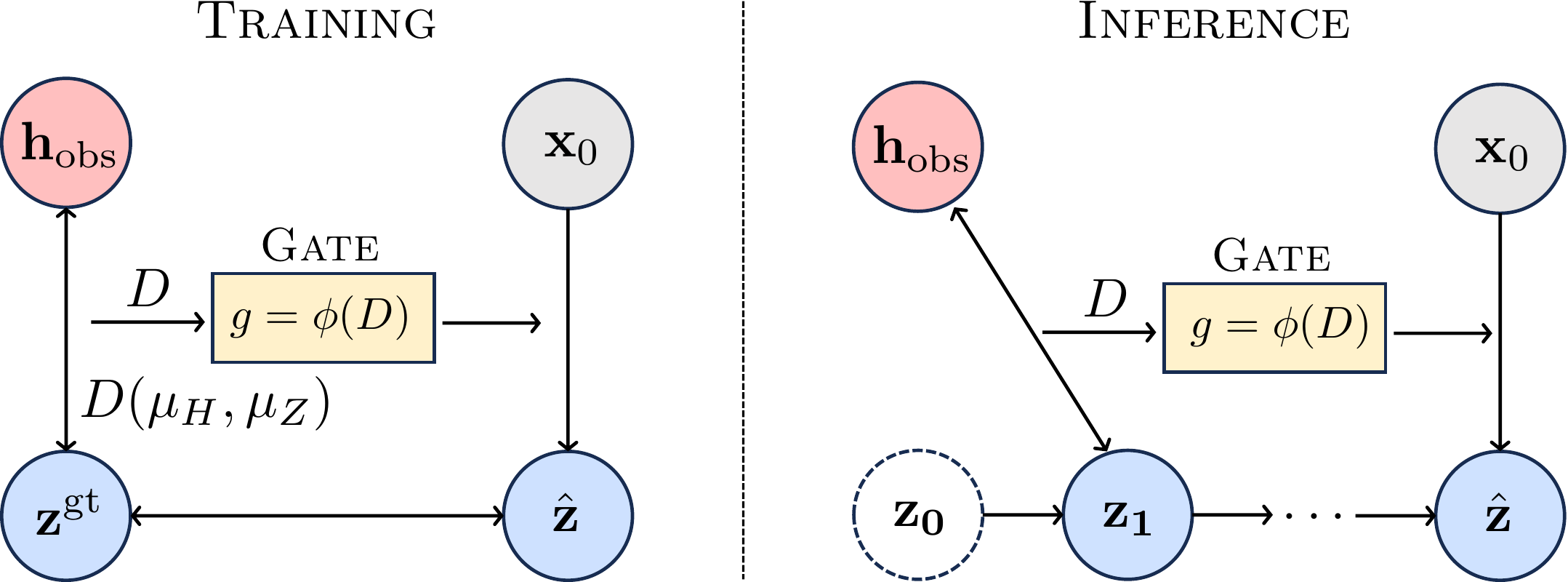}
  \caption{\textbf{Training and \DiGRefine.}
  During training, the gate is computed from ground-truth actions; $\sg(\cdot)$ marks the stop-gradient path, while $f$ remains on the normal action-generation path.
  During inference (\DiGRefine), the gate is computed from predicted actions and used inside the refinement loop.}
  \label{fig:train-infer}
\end{figure}

\subsection{Inference-time refinement with \DiGRefine}
\label{sec:dig-refine}

\DiGRefine is an inference-time procedure that reuses the same gate mapping and residual operator, but computes the action representation from predicted action chunks instead of ground-truth ones.

At control step $t$, the previous predicted chunk $\hat a_{t-1}$ is available.
This breaks the circular dependency between the current gate and the current prediction.
The current observation $o_t$ yields context features $H_t$.
An initial gate is computed from the current features and the previous chunk:
\begin{equation}
g_t^{\mathrm{prev}} \;=\; \phi\!\left(D(\mu_{H_t},\mu_{Z(\hat a_{t-1})})\right),
\qquad
H_t^{(0)} \;=\; H_t + \lambda\, g_t^{\mathrm{prev}}\,\cR(H_t),
\end{equation}
and the action expert produces an initial chunk $\hat a_t^{(0)}=\mathrm{FMHead}(H_t^{(0)})$. Further refinement uses the current prediction to update the chunk:
\begin{equation}
\hat a_t^{(i+1)} \;=\; \mathrm{FMHead}\!\left(H_t + \lambda\,\phi\!\left(D(\mu_{H_t},\mu_{Z(\hat a_t^{(i)})})\right)\cR(H_t)\right).
\label{eq:refine}
\end{equation}
After $N$ refinements, the policy executes the first action in $\hat a_t^{(N)}$ and replans at the next control step.
$N=0$ yields the single-pass version that relies only on $g_t^{\mathrm{prev}}$.
The loop is trust-gated rather than error-magnitude-gated.
Large discrepancy lowers $g$ and keeps the conditioned features close to the backbone baseline, instead of asking the residual branch to make a larger correction from an unreliable action representation.
As the predicted chunk becomes more compatible with the observation, the recomputed gate can increase and allow the learned residual to contribute more.
We do not assume a formal contraction theorem for this loop, since such a bound would require architecture-specific Lipschitz constants for the action expert.
The useful property is damping: high-discrepancy states remain conservative, while low-discrepancy states can use the residual path more strongly.
Section~\ref{sec:ablations} shows that a small number of rounds captures most gains.

\noindent\textbf{Scope of the single-centroid signal.}
The centroid summarizes the action chunk only for the reliability branch.
It is appropriate when the main question is whether the observation features and the action expert's internal coordinates are compatible at the chunk level.
It can be less discriminative when two distinct temporal modes have similar centroids, or when the failure is geometric or contact-level rather than representational.
Natural extensions are multi-centroid gates $\min_j D(\mu_H,\mu_{Z_j})$ and temporal-bin discrepancies $\sum_b D(\mu_{H_b},\mu_{Z_b})$.
If the action projector $f$ collapsed all chunks to a similar region, $D$ would also lose discriminative power; the stop-gradient design prevents directly optimizing such collapse through the gate, and the linear projector's limited capacity makes this failure unlikely in the reported settings.

\noindent\textbf{Reference pseudo-code.}
Algorithms~\ref{alg:train} and~\ref{alg:refine} summarize the training and inference procedures.

\begin{algorithm}[ht]
\caption{\DiG Training (one gradient step)}
\label{alg:train}
\begin{algorithmic}[1]
\Require Observation $o$, instruction $l$, demonstrated chunk $a_{1:K}$
\State Compute context tokens $H=\mathrm{Backbone}(o,l)$
\State Project actions $Z=f(a_{1:K})$ and compute $\bar z=\frac{1}{K}\sum_k z_k$
\State $D\gets \frac{1}{M}\sum_{m=1}^{M}\frac{1}{T}\sum_{i=1}^{T}(\theta_m^\top h_i-\theta_m^\top\bar z)^2$
\State $g\gets \max\{g_{\min},\exp(-\tau D)\}$
\State $\tilde H\gets H+\lambda\,g\,\cR(H)$
\State Sample $t\sim\mathrm{Uniform}[0,1]$ and form the interpolant $x_t$
\State $\cL\gets \sg(g)\,\cL_{\mathrm{FM}}(\tilde H,x_t,a_{1:K})$
\State Update the backbone, $\cR$, and action expert using $\nabla\cL$
\end{algorithmic}
\end{algorithm}

\begin{algorithm}[ht]
\caption{\DiGRefine Inference}
\label{alg:refine}
\begin{algorithmic}[1]
\Require Observation $o$, instruction $l$, previous chunk $\hat a_{\mathrm{prev}}$, refinement steps $N$
\State Compute $H=\mathrm{Backbone}(o,l)$ and cache $\cR(H)$
\State $\bar z^{\mathrm{prev}}\gets \frac{1}{K}\sum_j f(\hat a_{\mathrm{prev}})_j$
\State $D^{\mathrm{prev}}\gets \frac{1}{M}\sum_m\frac{1}{T}\sum_i(\theta_m^\top h_i-\theta_m^\top\bar z^{\mathrm{prev}})^2$
\State $g^{\mathrm{prev}}\gets \max\{g_{\min},\exp(-\tau D^{\mathrm{prev}})\}$
\State $H^{(0)}\gets H+\lambda\,g^{\mathrm{prev}}\,\cR(H)$; $\hat a^{(0)}\gets\mathrm{ActionExpert}(H^{(0)})$
\For{$k=0,\ldots,N-1$}
    \State $\bar z^{(k)}\gets \frac{1}{K}\sum_j f(\hat a^{(k)})_j$
    \State $D^{(k)}\gets \frac{1}{M}\sum_m\frac{1}{T}\sum_i(\theta_m^\top h_i-\theta_m^\top\bar z^{(k)})^2$
    \State $g^{(k)}\gets \max\{g_{\min},\exp(-\tau D^{(k)})\}$
    \State $H^{(k+1)}\gets H+\lambda\,g^{(k)}\,\cR(H)$
    \State $\hat a^{(k+1)}\gets\mathrm{ActionExpert}(H^{(k+1)})$
\EndFor
\State \Return the first action of $\hat a^{(N)}$ and replan at the next control step
\end{algorithmic}
\end{algorithm}

The single-pass setting corresponds to $N=0$, which uses only the previous-chunk gate $g^{\mathrm{prev}}$ and skips current-step refinement.
All reported experiments use $N=3$ for \DiGRefine.
Figure~\ref{fig:refine-ablation} shows that this setting is already on the early plateau.

\subsection{Computational overhead}
\label{sec:compute}

The \DiG module is designed to add negligible cost relative to the backbone and action expert.
The only learned component is the residual operator $\cR$, a single linear map $\R^d\!\to\!\R^d$.
This adds $d^2+d$ parameters.
For $d\!=\!1024$ the count is roughly 1M, less than 0.1\% of either backbone:
\begin{equation}
|\theta_{\mathrm{DiG}}| = d^2 + d, \qquad
\frac{|\theta_{\mathrm{DiG}}|}{|\theta_{\mathrm{backbone}}|} < 0.1\%.
\end{equation}
Training overhead consists of the discrepancy computation at $O(TM)$ and the residual forward pass at $O(Td^2)$.
Both are dominated by the backbone forward pass and the action expert's flow-matching iterations, so wall-clock training time increases by less than 3\% on 8$\times$A800 GPUs.
Total training time is approximately 10 hours for LIBERO and 20 hours for RoboCasa.
At inference time, the backbone is run once and $H$, $\cR(H)$ are cached.
Each refinement iteration requires only one action-expert forward pass.
With $N\!=\!3$ iterations the total refinement overhead is well within the budget of a 10\,Hz control loop.
Memory overhead is limited to caching $H\!\in\!\R^{T\times d}$ and $\cR(H)\!\in\!\R^{T\times d}$, which is negligible compared to the backbone activations.

\section{Experiments}
\label{sec:exp}

Experiments evaluate \DiG along three axes:
overall performance on simulation and real-robot benchmarks, robustness under distribution shift, and the contribution of each design component.
Two backbones are used to test compatibility with different architectures.
The first is $\pi_{0.5}$ \cite{intelligence2025pi05} with an action expert coupled to a large vision-language backbone.
The second is GR00T-N1 \cite{bjorck2025gr00t}, a dual-system VLA whose DiT action expert is conditioned on vision-language token embeddings.
In both cases, \DiG attaches at the representation interface consumed by the action expert.

\subsection{Training setup}
\label{sec:training-setup}

Both backbones are fine-tuned with their official recipes.
$\pi_{0.5}$ uses its default configuration, with the pretrained vision-language backbone frozen during warmup and then jointly trained.
GR00T-N1 follows its two-stage recipe, where the vision-language encoder is frozen and only the action model is updated.
The \DiG-specific settings are shared across all benchmarks and backbones; detailed values are listed in Appendix~\ref{app:hyperparams}.

\subsection{Simulation benchmarks}
\label{sec:sim}

Two simulation benchmarks are used.
\textbf{LIBERO} \cite{liu2023libero} is a 40-task suite for tabletop manipulation conditioned on language instructions, organized into four categories of increasing difficulty (Spatial, Object, Goal, Long).
\textbf{RoboCasa} \cite{nasiriany2024robocasa} is a photorealistic household simulator with 24 kitchen tasks spanning pick-and-place, door/drawer manipulation, and multi-step cooking.
RoboCasa is evaluated in a low-data regime with 50 demonstrations per task to stress-test generalization under limited supervision.

\noindent\textbf{Simulation task setting and evaluation protocol.}
LIBERO is reported with its standard 40-task split, with 10 tasks in each of the Spatial, Object, Goal, and Long categories.
These categories are useful to read as task families rather than as arbitrary names: Spatial tasks emphasize relative placement, Object tasks vary which object must be selected under a similar motion pattern, Goal tasks vary the desired terminal arrangement or receptacle, and Long tasks compose several primitives into multi-stage episodes.
Each LIBERO task is trained from 50 demonstrations and evaluated with 50 rollouts, so Table~\ref{tab:libero} aggregates 2000 evaluation rollouts per method.
Category scores are averaged over tasks within the split, and the overall average is the mean over all 40 tasks.

RoboCasa is reported on 24 kitchen tasks in the low-data regime with 50 demonstrations per task.
The three groups in Table~\ref{tab:robocasa} correspond to transfers into receptacles, articulated-object interactions such as opening or closing drawers and cabinets, and other multi-step kitchen behaviors that combine both.
Each RoboCasa task is evaluated with 50 rollouts, so the reported averages summarize 1200 evaluation rollouts per method.

\noindent\textbf{LIBERO observation and action spaces.}
Each LIBERO task provides a single $224\!\times\!224$ RGB image together with a proprioceptive state vector that encodes joint positions and the gripper opening:
\begin{equation}
o = \bigl(I_{\mathrm{RGB}} \in \R^{224\times224\times3},\; s_{\mathrm{prop}} \in \R^{d_s}\bigr).
\end{equation}
The action space is 7-dimensional, consisting of a 3-DoF position delta, a 3-DoF rotation delta, and a 1-DoF gripper command:
\begin{equation}
a = (\Delta p,\, \Delta r,\, g_{\mathrm{grip}}) \in \R^{7}.
\end{equation}
The chunk size $K$ is inherited from the backbone.

\noindent\textbf{RoboCasa observation and action spaces.}
RoboCasa shares the same observation and action representation as LIBERO: a single $224\!\times\!224$ RGB image, a proprioceptive state vector, and 7-dimensional end-effector actions.
The key difference is the data regime: each task uses only 50 demonstrations, making it a low-data benchmark that stresses generalization from limited coverage.

\noindent\textbf{Receding-horizon execution.}
All benchmarks use chunked receding-horizon control.
At each control step, the policy predicts an action chunk, executes only the first action, receives a new observation, and replans.
This is the regime in which the discrepancy gate is most informative, because the gate is recomputed at every decision point rather than once per episode.

\begin{table}[tb]
\centering
\caption{\textbf{Success rates (\%) on LIBERO.}
Mean success over 50 rollouts per task.}
\label{tab:libero}
\setlength{\tabcolsep}{8pt}
\resizebox{.72\linewidth}{!}{
\begin{tabular}{lccccc}
\toprule
\textbf{Method} & \textbf{Spatial} & \textbf{Object} & \textbf{Goal} & \textbf{Long} & \textbf{Avg} \\
\midrule
Diffusion Policy \cite{chi2023dp} & 78.5 & 87.5 & 73.5 & 64.8 & 76.1 \\
SpatialVLA \cite{qu2024spatialvla} & 88.2 & 89.9 & 78.6 & 55.5 & 78.1 \\
CoT-VLA \cite{zhao2025cotvla} & 87.5 & 91.6 & 87.6 & 69.0 & 83.9 \\
OpenVLA \cite{kim2024openvla} & 84.7 & 88.4 & 79.2 & 53.7 & 76.5 \\
OpenVLA-OFT \cite{kim2025fine} & 97.6 & 98.4 & 97.9 & 94.5 & 97.1 \\
GR00T-N1 \cite{bjorck2025gr00t} & 94.4 & 97.6 & 93.0 & 90.6 & 93.9 \\
$\pi_0$ \cite{black2024pi0} & 98.0 & 96.8 & 94.4 & 88.4 & 94.4 \\
$\pi_{0.5}$ \cite{intelligence2025pi05} & 98.8 & 98.2 & 98.0 & 92.4 & 96.9 \\
\midrule
\rowcolor{BlockA!30}GR00T-N1+\DiG (Ours) & 96.0 & 98.4 & 94.8 & 92.1 & 95.3 \\
\rowcolor{BlockA!30}$\pi_{0.5}$+\DiG (Ours) & \textbf{99.2} & \textbf{99.0} & \textbf{98.6} & \textbf{96.4} & \textbf{98.3} \\
\bottomrule
\end{tabular}
}
\end{table}

\begin{table}[t]
\centering
\caption{\textbf{RoboCasa 24-task benchmark with 50 demonstrations per task.}
Success rates (\%) averaged within categories over 50 rollouts per task.}
\label{tab:robocasa}
\setlength{\tabcolsep}{8pt}
\resizebox{.8\linewidth}{!}{
\begin{tabular}{lcccc}
\toprule
\textbf{Method} & \textbf{Pick \& Place} & \textbf{Doors/Drawers} & \textbf{Others} & \textbf{Avg} \\
\midrule
GR00T-N1 & 18.6 & 50.2 & 39.1 & 36.0 \\
\rowcolor{BlockA!30}GR00T-N1+\DiG (Ours) & 22.4 & 60.3 & 47.0 & \textbf{43.2} \\
\midrule
$\pi_{0.5}$ & 21.5 & 57.8 & 44.9 & 41.4 \\
\rowcolor{BlockA!30}$\pi_{0.5}$+\DiG (Ours) & 27.2 & 73.4 & 57.2 & \textbf{52.6} \\
\bottomrule
\end{tabular}
}
\end{table}

\noindent\textbf{Main results on LIBERO.}
Table~\ref{tab:libero} compares \DiG against recent VLA baselines.
\DiG improves both backbones and achieves the highest average success rate.
The gains are not uniform across task categories: LIBERO-Spatial and LIBERO-Object, where the baseline already exceeds 98\%, see modest improvements ($+$0.4 to $+$0.8), while LIBERO-Long improves by $+$4.0 with $\pi_{0.5}$.
This asymmetry is expected: long-horizon tasks involve more control steps, so early representational errors compound and the reliability gate has more opportunities to intervene.
On shorter tasks where the baseline is near ceiling, the gate values stay high and \DiG reduces to near-identity refinement. GR00T-N1+\DiG also improves over the GR00T-N1 baseline ($+$1.4 avg), confirming that the module generalizes across architectures with different action-head designs.

\noindent\textbf{Few-shot results on RoboCasa.}
Table~\ref{tab:robocasa} shows that \DiG yields larger absolute gains here than on LIBERO ($+$11.2 avg for $\pi_{0.5}$, $+$7.2 for GR00T-N1).
With scarce demonstrations, the backbone is more prone to overfitting to spurious correlations in the training set.
The transport discrepancy acts as a soft regularizer: samples whose observation features are poorly aligned with the action representation receive lower gate values, reducing their influence on the loss.
The effect is strongest on Pick \& Place ($+$5.7 for $\pi_{0.5}$) and Doors/Drawers ($+$15.6), categories that involve spatial alignment where spurious visual cues are most harmful.
\begin{table}[tb]
\centering
\caption{\textbf{Robustness to non-stationary perturbations.}
Success rates (\%) under time-varying noise.
Numbers in color show change relative to the baseline row.}
\label{tab:nonstat}
\setlength{\tabcolsep}{8pt}
\resizebox{.85\linewidth}{!}{
\begin{tabular}{llccccc}
\toprule
\textbf{Perturbation} & \textbf{Method} & \textbf{Spatial} & \textbf{Object} & \textbf{Goal} & \textbf{Long} & \textbf{Avg} \\
\midrule
\multirow{2}{*}{Cosine only}
&  w/o \DiG & \textbf{87.6} & \textbf{89.8} & 86.4 & 70.2 & 83.5 \\
&  w/ \DiG & 87.4 {\color{red}\scriptsize{-0.2}} & 89.8 {\color{black}\scriptsize{0.0}} & \textbf{91.6} {\color{green!60!black}\scriptsize{+5.2}} & \textbf{80.0} {\color{green!60!black}\scriptsize{+9.8}} & \textbf{87.2} {\color{green!60!black}\scriptsize{+3.7}} \\
\midrule
\multirow{2}{*}{Sine only}
&  w/o \DiG & 88.2 & 88.4 & 87.2 & 68.6 & 83.1 \\
&  w/ \DiG & \textbf{90.0} {\color{green!60!black}\scriptsize{+1.8}} & \textbf{90.6} {\color{green!60!black}\scriptsize{+2.2}} & \textbf{91.2} {\color{green!60!black}\scriptsize{+4.0}} & \textbf{80.4} {\color{green!60!black}\scriptsize{+11.8}} & \textbf{88.1} {\color{green!60!black}\scriptsize{+5.0}} \\
\midrule
\multirow{2}{*}{Cosine + sine}
&  w/o \DiG & \textbf{87.8} & 87.2 & 86.0 & 67.8 & 82.2 \\
&  w/ \DiG & 86.4 {\color{red}\scriptsize{-1.4}} & \textbf{88.2} {\color{green!60!black}\scriptsize{+1.0}} & \textbf{90.8} {\color{green!60!black}\scriptsize{+4.8}} & \textbf{79.2} {\color{green!60!black}\scriptsize{+11.4}} & \textbf{86.2} {\color{green!60!black}\scriptsize{+4.0}} \\
\bottomrule
\end{tabular}
}
\end{table}

\noindent\textbf{Non-stationary perturbations.}
Table~\ref{tab:nonstat} adds time-varying noise (cosine, sine, or both) to observations and proprioception following the protocol of \cite{feng2022factored,zhang2024tackling}.
A consistent pattern emerges across all perturbation types: \DiG improves substantially on hard tasks like Goal and Long ($+$4.0 to $+$11.8).
Spatial tasks are short and the perturbation has limited time to accumulate drift.
Goal and Long tasks require sustained multi-step execution where time-varying noise compounds through the chunked control loop.

\subsection{Real robot benchmarks}
\label{sec:real}

\begin{figure}[tb]
  \centering
  \includegraphics[width=\linewidth]{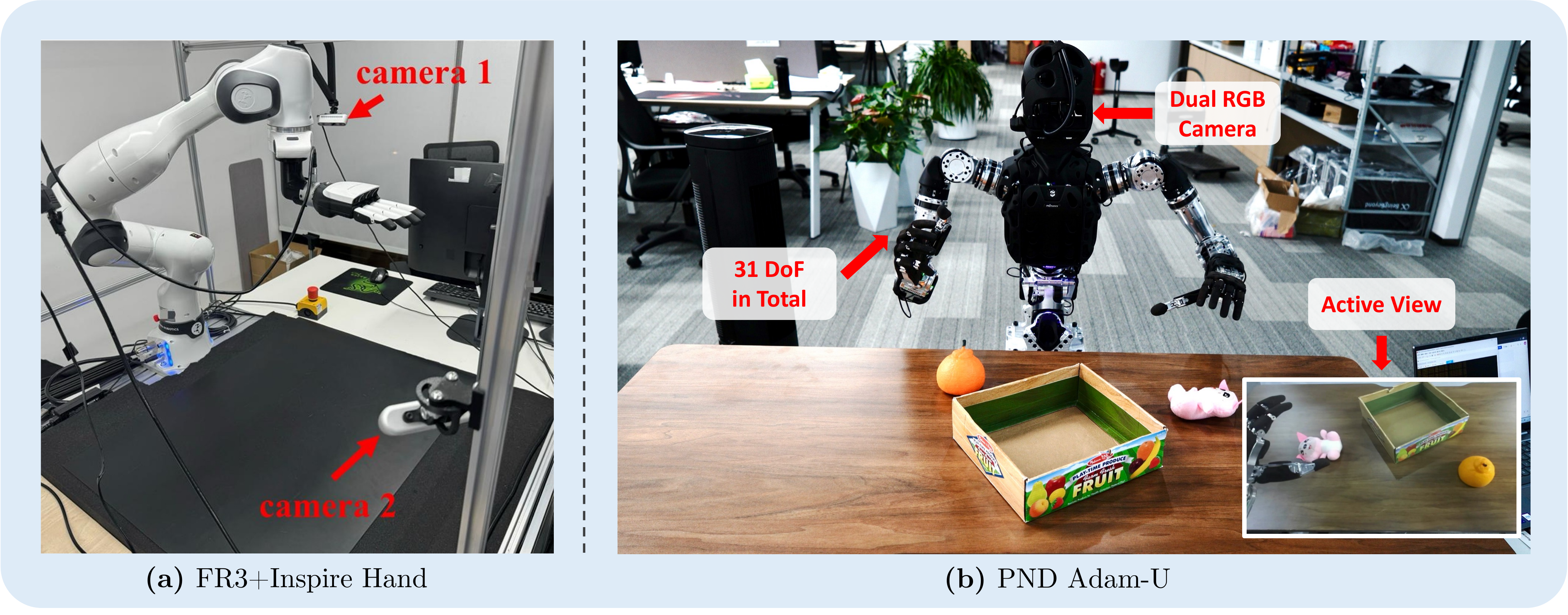}
  \caption{\textbf{(a) FR3+Inspire Hand Setup.}
  Two cameras and a dexterous hand.
  \textbf{(b) PND Adam-U Setup.}
  A 31-DoF humanoid with dual dexterous hands and an actively controlled head mounting dual cameras.}
  \label{fig:real-setup}
\end{figure}

Simulation results establish that \DiG improves performance in controlled settings.
Real-robot evaluation tests whether these gains persist under real sensor noise, contact dynamics, and human intervention.
Two platforms are used (Figure~\ref{fig:real-setup}):
(i)~a 7-DoF Franka FR3 arm equipped with an Inspire dexterous hand, observed by two fixed RGB cameras, evaluated on four tabletop manipulation tasks with 20 trials per task;
(ii)~a PND Adam-U humanoid upper body (31 DoF) with dual dexterous hands and an actively controlled head mounting dual RGB cameras, evaluated on an active-view clean-up task with 20 trials per configuration.
Both platforms use $\pi_{0.5}$ as the backbone.
We first specify the control loops and perturbation settings, then report success rates.

\noindent\textbf{Real-robot protocol: Franka FR3 + Inspire Hand.}
Figure~\ref{fig:real-setup}(a) shows the Franka setup. Table~\ref{tab:hardware} summarizes the hardware configuration of both platforms.
The platform uses a 7-DoF FR3 arm with a 6-DoF Inspire dexterous hand, two fixed RGB cameras that view the tabletop from complementary angles, and proprioceptive arm and hand state.
The policy is queried at 10\,Hz and produces a 13-dimensional command spanning the 7 arm DoF and 6 hand DoF.
On Franka, each 10\,Hz policy output is executed by a 50\,Hz interpolated low-level controller.
Each episode uses the same receding-horizon execution as the simulation benchmarks.
Across the four Franka tasks, 20 in-distribution trials and 20 OOD trials are collected per task, so Table~\ref{tab:franka} summarizes 160 real episodes per method.

\begin{table}[t]
\centering
\caption{Real-robot hardware specifications. DoF counts refer to the policy-controlled dimensions.
}
\label{tab:hardware}
\small
\begin{tabular}{lccccc}
\toprule
\textbf{Platform} & \textbf{Controlled DoF} & \textbf{Cameras} & \textbf{Action dim} & \textbf{Policy rate} & \textbf{Execution} \\
\midrule
FR3 + Inspire & 7 arm + 6 hand & 2$\times$ fixed RGB & 13 & 10\,Hz & 50\,Hz interp. \\
Adam-U & 31 upper-body & 2$\times$ head RGB & 31 & 10\,Hz & 10\,Hz \\
\bottomrule
\end{tabular}
\end{table}

\begin{figure}[tb]
  \centering
  \includegraphics[width=\linewidth]{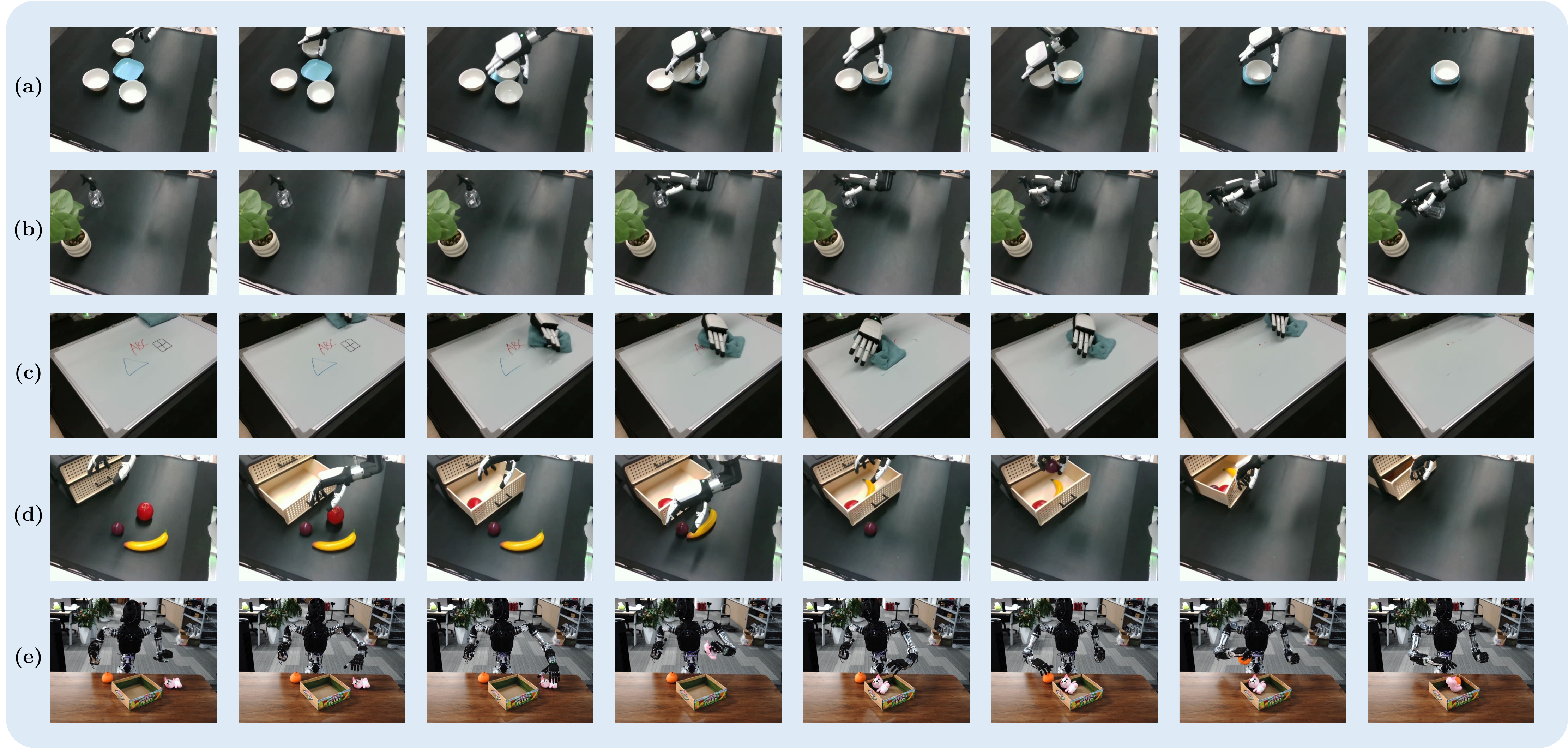}
  \caption{\textbf{Qualitative rollouts on the real robots.}
  Rows (a) to (d) are the four Franka tasks and row (e) is the Adam-U active-view clean-up task.
  Each row shows one representative successful episode from left to right. Videos are provided as ancillary material.}
  \label{fig:real-exp}
\end{figure}

The four Franka tasks are intentionally heterogeneous.
\textbf{Stack-Bowls} is a sequential placement task where the robot must lift bowls one by one and place them onto the target stack without disturbing previously placed objects.
\textbf{Spray-Plant} is a tool-use task where the robot must grasp the bottle with the correct orientation, transport it toward the plant, and trigger spraying near the leaves.
\textbf{Wipe-Whiteboard} is a contact-rich task where the robot must reach the cloth or eraser and then remove the marker strokes on the board with sustained surface contact.
\textbf{Sort-Into-Drawer} is the longest-horizon setting: several objects must be picked in sequence and placed into the correct drawer organizer compartments.
These four cases correspond to rows (a) to (d) of Figure~\ref{fig:real-exp}.
The sub-task score averages completion of the intermediate stages of each episode, while the whole-task score requires that every stage be completed in order within one trial.

\begin{table}[tb]
\centering
\caption{\textbf{Franka real-robot success rates (\%).}
We report sub-task / whole-task success over 20 trials.
}
\label{tab:franka}
\setlength{\tabcolsep}{4pt}
\resizebox{.8\linewidth}{!}{
\begin{tabular}{llcccc}
\toprule
& \textbf{Method} & \textbf{Stack-Bowls} & \textbf{Spray-Plant} & \textbf{Wipe-Whiteboard} & \textbf{Sort-Into-Drawer} \\
\midrule
\multirow{2}{*}{\textbf{In-dist}}
& $\pi_{0.5}$ & 60 / 40 & 65 / 45 & 60 / 40 & 50 / 35 \\
& \cellcolor{BlockA!30}$\pi_{0.5}$+\DiG (Ours) & \cellcolor{BlockA!30}\textbf{65} / \textbf{50} & \cellcolor{BlockA!30}\textbf{70} / \textbf{50} & \cellcolor{BlockA!30}\textbf{70} / \textbf{50} & \cellcolor{BlockA!30}\textbf{60} / \textbf{40} \\
\midrule
\multirow{2}{*}{\textbf{OOD}}
& $\pi_{0.5}$ & 40 / 15 & 35 / 10 & 45 / 20 & 50 / 20 \\
& \cellcolor{BlockA!30}$\pi_{0.5}$+\DiG (Ours) & \cellcolor{BlockA!30}\textbf{65} / \textbf{40} & \cellcolor{BlockA!30}\textbf{60} / \textbf{30} & \cellcolor{BlockA!30}\textbf{60} / \textbf{30} & \cellcolor{BlockA!30}\textbf{60} / \textbf{35} \\
\bottomrule
\end{tabular}
}
\end{table}

\begin{figure}[tb]
  \centering
  \includegraphics[width=\linewidth]{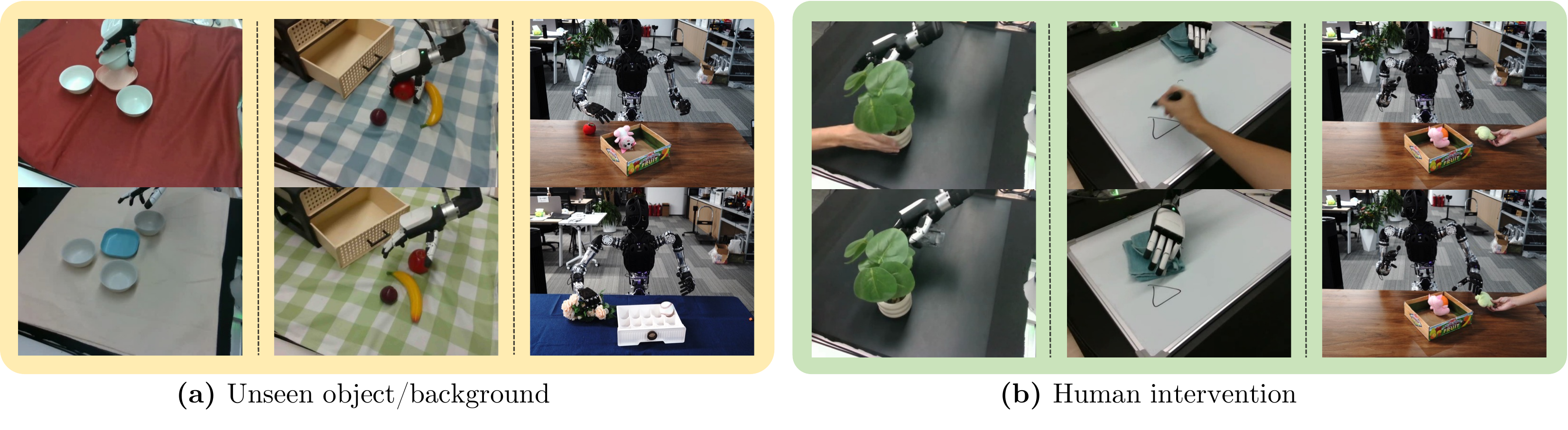}
  \caption{\textbf{Perturbations in the OOD setting.}
  Background textures are changed and human intervention is introduced during execution.}
  \label{fig:real-robot-perturb}
\end{figure}

The OOD protocol is task dependent.
Appearance and background changes are used for the layout-sensitive placement tasks.
Human intervention during execution is used for the non-stationary tasks.
In practice, the first group is Stack-Bowls and Sort-Into-Drawer, and the second group is Spray-Plant and Wipe-Whiteboard.
For clarity, Spray-Plant belongs to the human-intervention setting rather than to the pure appearance-shift setting.
Figure~\ref{fig:real-robot-perturb} illustrates these perturbations.
The intent is to probe two failure modes separately: visual appearance shift and online scene change after the policy has already committed to a chunk.

\begin{figure}[tb]
  \centering
  \includegraphics[width=\linewidth]{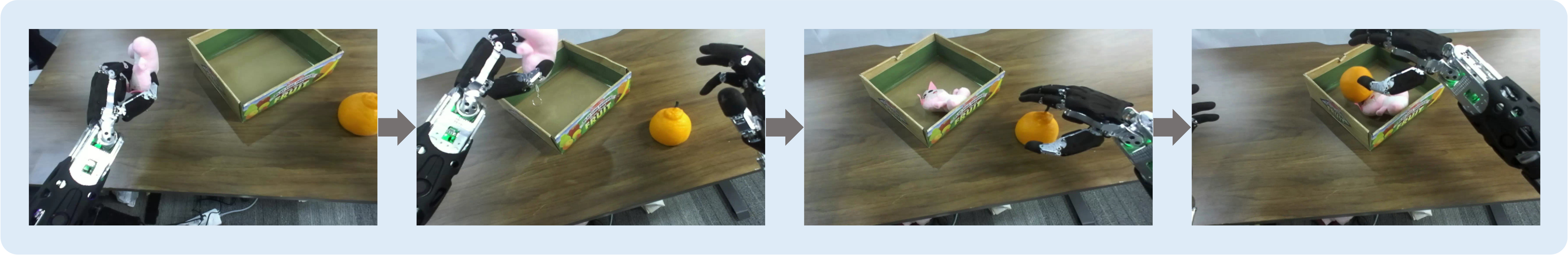}
  \caption{\textbf{Adam-U active-view evaluation.}
  The head camera is actively controlled during execution.
  The seen and unseen settings differ in background and lighting.}
  \label{fig:adamu-view}
\end{figure}

\noindent\textbf{Real-robot protocol: PND Adam-U humanoid.}
Figures~\ref{fig:real-setup}(b) and~\ref{fig:adamu-view} show the Adam-U setup.
The policy-controlled upper body has 31 DoF in total, including both arms, both hands, and the actively controlled head.
Observations consist of the dual head-camera RGB views together with proprioceptive joint state.
The policy outputs upper-body joint targets at 10\,Hz.
The benchmark is an active-view clean-up task in which 1 to 3 loose objects are collected into a box while the head camera is actively repositioned to keep the workspace in view.
The number of objects controls the effective horizon length: 1 object is a short reach-and-place episode, while 2 and 3 objects require repeated re-centering of the scene after each pickup.
Each object-count setting is evaluated for 20 trials under seen backgrounds and 20 under unseen backgrounds, so the left block of Table~\ref{tab:adamu} summarizes 120 real episodes per method.
The seen and unseen conditions keep the task logic fixed but change the background cloth, box appearance, and lighting.

\noindent\textbf{Franka results.}
Table~\ref{tab:franka} reports sub-task and whole-task success rates across four tasks.
In the in-distribution setting, \DiG improves all tasks, with whole-task gains ranging from $+$5 (Spray-Plant, Sort-Into-Drawer) to $+$10 (Stack-Bowls, Wipe-Whiteboard).
Under OOD conditions (background texture changes and human intervention, visualized in Figure~\ref{fig:real-robot-perturb}), the gap widens substantially.
The baseline whole-task success drops to 10\% to 20\%, while \DiG maintains 30\% to 40\%.
The largest relative improvement appears on Spray-Plant under OOD ($+$25 sub-task, $+$20 whole-task), a tool-use task where visual distractors can cause the policy to mislocate the bottle.
Sort-Into-Drawer, the longest-horizon task, shows the smallest in-distribution gap, yet \DiG still provides meaningful improvement under OOD, consistent with the pattern observed in simulation.

\noindent\textbf{Adam-U and visual-shift results.}
Table~\ref{tab:adamu} evaluates a 31-DoF humanoid on an active-view clean-up task with 1 to 3 objects under seen and unseen backgrounds (Figure~\ref{fig:adamu-view}).
In the seen setting, \DiG matches or improves the baseline, with gains growing as the number of objects increases ($+$0 for 1 object, $+$10 for 3 objects).
More objects require longer execution and more head-camera viewpoint changes, amplifying the chance of representational drift.
Under unseen backgrounds, the baseline drops sharply (75 $\to$ 55 for 1 object, 35 $\to$ 25 for 3 objects).
\DiG recovers a substantial fraction: the 3-object unseen success rate improves from 25\% to 40\%, a $+$15 absolute gain.
This confirms that the transport discrepancy captures background-induced drift that is invisible to the flow-matching loss alone.

\begin{table}[tb]
\centering
\caption{\textbf{Adam-U and visual-shift real-task success rates (\%).}
The left block reports active-view clean-up with 1 to 3 objects under seen and unseen background.
The right block reports visual-shift tasks under standard and OOD settings; AF: Arrange-Flower, BH: Bimanual Handover.
20 trials per cell.
}
\label{tab:adamu}
\setlength{\tabcolsep}{5pt}
\resizebox{.8\linewidth}{!}{
\begin{tabular}{lcccccccccc}
\toprule
\multirow{3}{*}{\textbf{Method}}
& \multicolumn{6}{c}{\textbf{Adam-U active-view clean-up}}
& \multicolumn{4}{c}{\textbf{Visual-shift real tasks}} \\
\cmidrule(lr){2-7} \cmidrule(lr){8-11}
& \multicolumn{3}{c}{\textbf{Seen}} & \multicolumn{3}{c}{\textbf{Unseen}}
& \multicolumn{2}{c}{\textbf{Std}} & \multicolumn{2}{c}{\textbf{OOD}} \\
\cmidrule(lr){2-4} \cmidrule(lr){5-7} \cmidrule(lr){8-9} \cmidrule(lr){10-11}
& \textbf{1 obj} & \textbf{2 objs} & \textbf{3 objs}
& \textbf{1 obj} & \textbf{2 objs} & \textbf{3 objs}
& \textbf{AF} & \textbf{BH} & \textbf{AF} & \textbf{BH} \\
\midrule
$\pi_{0.5}$ & 75 & 60 & 35 & 55 & 30 & 25 & 15 & 35 & 5 & 15 \\
\rowcolor{BlockA!30}$\pi_{0.5}$+\DiG (Ours) & \textbf{75} & \textbf{65} & \textbf{45} & \textbf{65} & \textbf{45} & \textbf{40} & \textbf{35} & \textbf{55} & \textbf{25} & \textbf{40} \\
\bottomrule
\end{tabular}
}
\end{table}

\begin{figure}[tb]
  \centering
  \includegraphics[width=\linewidth]{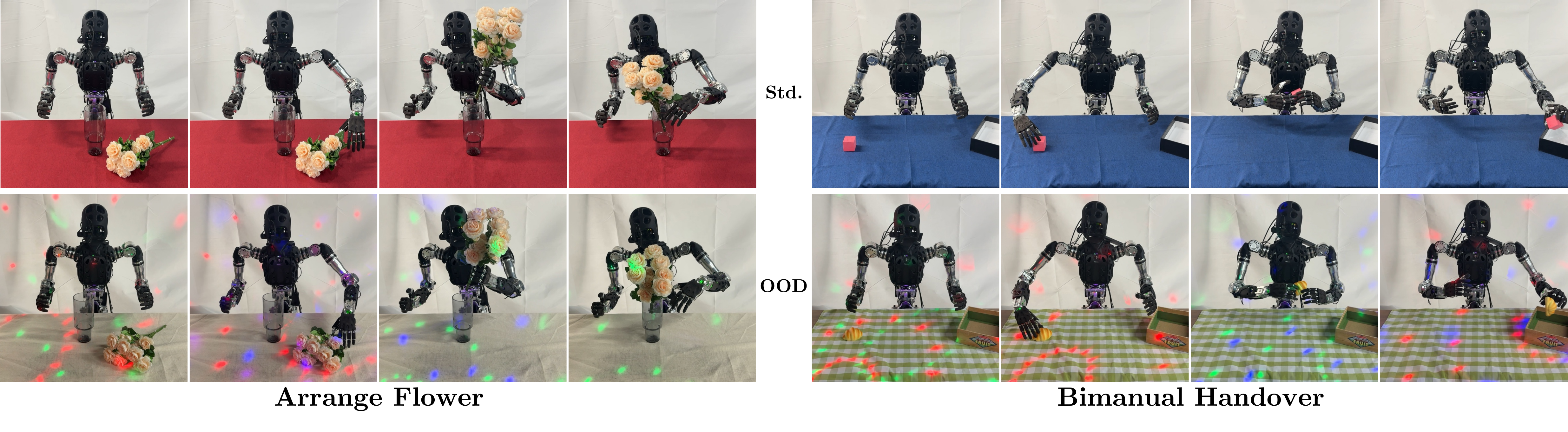}
  \caption{\textbf{Real visual-shift settings.}
  Standard and OOD views for Arrange-Flower and Bimanual Handover.}
  \label{fig:hard-real-shift}
\end{figure}

\noindent\textbf{Visual-shift real tasks.}
The right block of Table~\ref{tab:adamu} and Figure~\ref{fig:hard-real-shift} evaluate two real tasks that keep the task semantics fixed while changing backgrounds and lighting.
Arrange-Flower stresses object rearrangement under cluttered visual context, and Bimanual Handover stresses coordinated dual-arm motion under the same appearance shift.
\DiG improves both standard and shifted variants, with larger gains in the OOD setting.

\noindent\textbf{Qualitative rollouts.}
Figure~\ref{fig:real-exp} samples representative successful rollouts from the real-robot tasks.
The qualitative view highlights where discrepancy-aware refinement matters: grasp-to-place transitions, contact onset, long-horizon scene changes, and active-view shifts.

\begin{itemize}
\item \textbf{Stack-Bowls (row a).} The key transition is the switch from grasping to placing, when the target stack moves into the central field of view and small placement errors can disturb previously stacked bowls. Discrepancy-aware refinement gives the policy an additional consistency check before executing this placement chunk.

\item \textbf{Spray-Plant (row b).} Successful spraying depends on both bottle orientation and transport to the plant. When human intervention changes the scene after the grasp, the previous action chunk can become stale; recomputing the gate with the current prediction helps the policy update the subsequent transport motion.

\item \textbf{Wipe-Whiteboard (row c).} This task is contact-rich, and the critical moment is contact onset. A misaligned approach angle leads to insufficient surface pressure, so a reliability signal at the observation-action interface is useful before the wiping motion enters a stable sliding phase.

\item \textbf{Sort-Into-Drawer (row d).} This task accumulates errors across several pick-and-place stages. After each object is moved, the scene changes for the next pick; recomputing discrepancy at every control step helps prevent an early grasp or placement error from propagating through the remaining sequence.

\item \textbf{Adam-U clean-up (row e).} The humanoid actively repositions its head cameras, producing large visual changes between consecutive control steps. \DiGRefine is particularly cheap in this setting because the backbone features are cached within one control step and only the action expert is rerun during refinement.
\end{itemize}

\noindent\textbf{Observed failure patterns.}
Two recurring failure modes are consistent with the mechanism described above.
First, the gate can occasionally assign low discrepancy to actions that are feature-level coherent but physically infeasible.
This occurs when the task geometry exceeds the expressiveness of the shared feature space, \eg grasping extremely thin objects whose contact constraints are not captured by the linear projector $f$.
Second, in long-horizon episodes, an early sub-task failure can shift the scene far from the support of all training demonstrations.
The gate can still indicate high discrepancy in this regime, but the residual operator's limited capacity (a single linear map) is insufficient to recover a viable action chunk from a fundamentally out-of-support observation.

\subsection{Ablations and analysis}
\label{sec:ablations}


\begin{table}[tb]
\centering
\caption{\textbf{Ablations on LIBERO (success \%).}
Metric columns vary the discrepancy with gate form fixed as $g=\exp(-\tau D)$.
Gate columns vary the gate with discrepancy fixed as sliced Wasserstein.
Red numbers denote drops from Ours.}
\label{tab:ablation-combined}
\setlength{\tabcolsep}{8pt}
\resizebox{.9\linewidth}{!}{
\begin{tabular}{lcccccccc}
\toprule
\multirow{2}{*}{\textbf{Task}}
& \multicolumn{3}{c}{\textbf{Metric ablation}}
& \multicolumn{4}{c}{\textbf{Gate ablation}}
& \multirow{2}{*}{\textbf{Ours}} \\
\cmidrule(lr){2-4}\cmidrule(lr){5-8}
& \textbf{Cos.} & \textbf{MMD} & \textbf{Sink.}
& \textbf{Fix.} & \textbf{Rand.} & \textbf{$\lambda=0$} & \textbf{MLP} &  \\
\midrule
\textbf{Spatial}
& 98.2 {\color{red}\scriptsize{-1.0}}
& 98.4 {\color{red}\scriptsize{-0.8}}
& 98.8 {\color{red}\scriptsize{-0.4}}
& 96.8 {\color{red}\scriptsize{-2.4}}
& 95.2 {\color{red}\scriptsize{-4.0}}
& 98.6 {\color{red}\scriptsize{-0.6}}
& 98.8 {\color{red}\scriptsize{-0.4}}
& \cellcolor{BlockA!30}\textbf{99.2} \\
\textbf{Object}
& 97.8 {\color{red}\scriptsize{-1.2}}
& 98.2 {\color{red}\scriptsize{-0.8}}
& 98.6 {\color{red}\scriptsize{-0.4}}
& 95.4 {\color{red}\scriptsize{-3.6}}
& 93.8 {\color{red}\scriptsize{-5.2}}
& 98.4 {\color{red}\scriptsize{-0.6}}
& 98.4 {\color{red}\scriptsize{-0.6}}
& \cellcolor{BlockA!30}\textbf{99.0} \\
\textbf{Goal}
& 97.2 {\color{red}\scriptsize{-1.4}}
& 97.6 {\color{red}\scriptsize{-1.0}}
& 98.2 {\color{red}\scriptsize{-0.4}}
& 92.2 {\color{red}\scriptsize{-6.4}}
& 89.4 {\color{red}\scriptsize{-9.2}}
& 98.0 {\color{red}\scriptsize{-0.6}}
& 98.0 {\color{red}\scriptsize{-0.6}}
& \cellcolor{BlockA!30}\textbf{98.6} \\
\textbf{Long}
& 92.6 {\color{red}\scriptsize{-3.8}}
& 93.4 {\color{red}\scriptsize{-3.0}}
& 95.8 {\color{red}\scriptsize{-0.6}}
& 84.6 {\color{red}\scriptsize{-11.8}}
& 80.8 {\color{red}\scriptsize{-15.6}}
& 94.2 {\color{red}\scriptsize{-2.2}}
& 94.8 {\color{red}\scriptsize{-1.6}}
& \cellcolor{BlockA!30}\textbf{96.4} \\
\textbf{Avg}
& 96.5 {\color{red}\scriptsize{-1.8}}
& 96.8 {\color{red}\scriptsize{-1.5}}
& 97.9 {\color{red}\scriptsize{-0.4}}
& 92.3 {\color{red}\scriptsize{-6.0}}
& 89.8 {\color{red}\scriptsize{-8.5}}
& 97.3 {\color{red}\scriptsize{-1.0}}
& 97.5 {\color{red}\scriptsize{-0.8}}
& \cellcolor{BlockA!30}\textbf{98.3} \\
\bottomrule
\end{tabular}}
\end{table}

\noindent\textbf{Discrepancy choice.}
The metric-ablation columns in Table~\ref{tab:ablation-combined} vary the discrepancy while holding the gate form fixed at $g=\exp(-\tau D)$.
All transport-based metrics outperform cosine distance and MMD, with sliced Wasserstein achieving the best average.
The gap is most visible on LIBERO-Long ($-$3.8 for cosine, $-$3.0 for MMD, $-$0.6 for Sinkhorn vs.\ sliced Wasserstein).
Cosine distance collapses the distributional structure into a single scalar similarity, losing information about how the feature distribution spreads around the action centroid.
MMD with an RBF kernel captures distributional differences but is sensitive to bandwidth selection; a single bandwidth cannot simultaneously resolve fine-grained structure in high-density regions and coarse drift in the tails.
Sinkhorn is closest to sliced Wasserstein in performance, which is expected since both are transport-based; the sliced variant is cheaper to compute and avoids the entropic regularization bias.
The performance gap between metrics widens on longer-horizon tasks.
On Spatial and Object, where episodes are short and the baseline is near ceiling, all four metrics produce similar gate distributions and the residual correction has limited room to help.
On Long, representational drift accumulates over many control steps, and the gate must distinguish between mild and severe misalignment.
Transport-based metrics preserve this graded information, while cosine distance saturates early and MMD bandwidth mismatches blur the distinction.

\noindent\textbf{Gate mechanism.}
The gate-ablation columns in Table~\ref{tab:ablation-combined} vary the gate while holding the discrepancy fixed at sliced Wasserstein.
A fixed gate ($g=0.5$) applies the same residual correction regardless of the input, removing all discrepancy-based modulation.
This ablation still benefits from the residual operator $\cR$ but loses the ability to adapt per sample.
The $-$6.0 avg drop (and $-$11.8 on Long) confirms that discrepancy-driven gating is the primary source of improvement, not the residual correction alone.
A random gate performs even worse ($-$8.5 avg), since random modulation introduces noise into the refinement and the loss reweighting.
Two targeted variants separate the roles of loss reweighting, residual refinement, and discrepancy estimation.
Setting $\lambda=0$ removes the residual branch but keeps loss reweighting, so the reliability signal alone improves over the baseline while remaining below full \DiG.
The MLP gate replaces the transport discrepancy with a learned scalar gate on pooled $H/\bar z$ while keeping the same residual and loss paths; its smaller gain indicates that the geometric discrepancy is more informative than an unconstrained learned gate.
The targeted variants close part of the gap but stay below full \DiG, while the non-adaptive controls follow fixed gate $>$ random gate.
The transport gate adds sample-adaptive modulation on top, and its advantage grows with task horizon, confirming that per-step reliability estimation is most valuable when errors compound.

\begin{figure}[tb]
  \centering
  \begin{subfigure}[t]{0.48\linewidth}
    \centering
    \includegraphics[width=\linewidth]{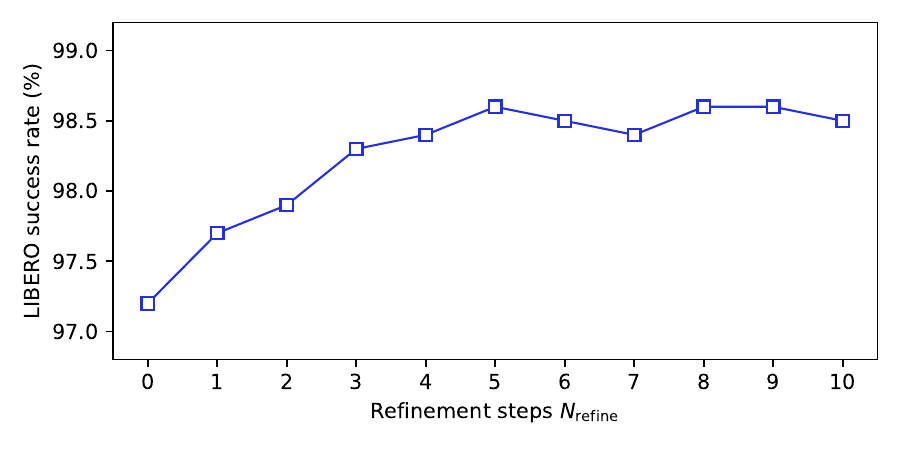}
    \caption{\textbf{Refinement iterations.}
    Success rate vs.\ refinement steps.}
    \label{fig:refine-ablation}
  \end{subfigure}
  \hfill
  \begin{subfigure}[t]{0.48\linewidth}
    \centering
    \includegraphics[width=\linewidth]{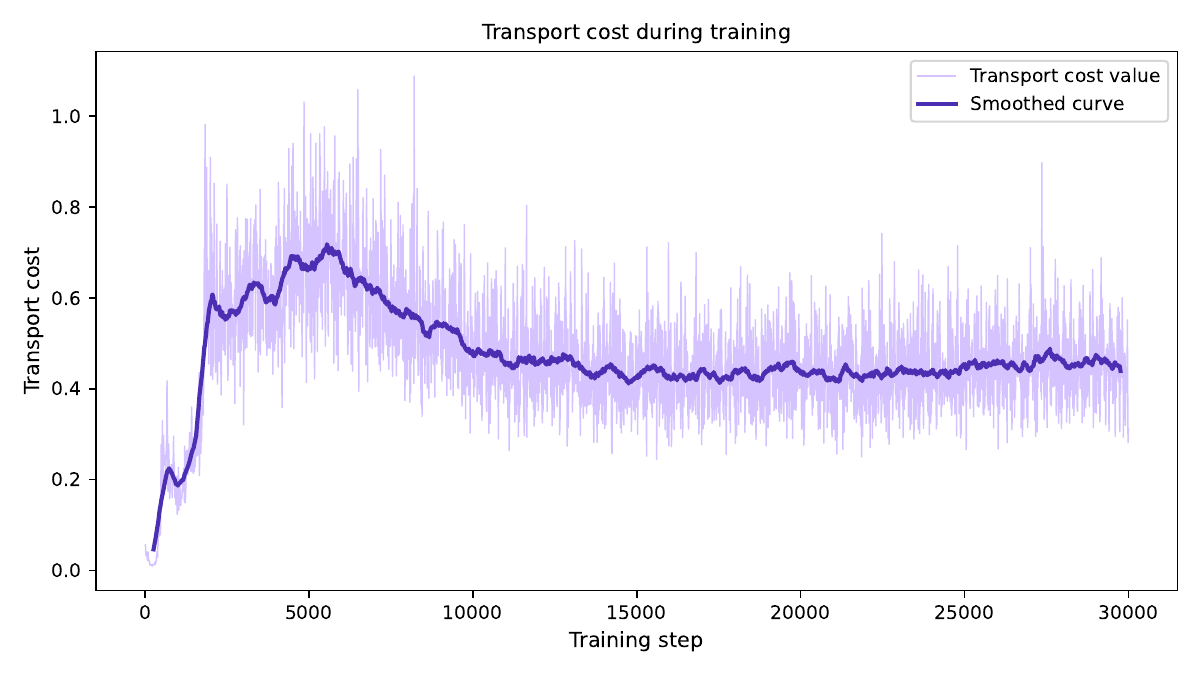}
    \caption{\textbf{Training discrepancy.}
    Raw and smoothed cost vs.\ training steps.}
    \label{fig:transport-curve}
  \end{subfigure}
  \caption{\textbf{Ablation diagnostics.}}
\end{figure}

\noindent\textbf{Refinement iterations.}
\DiGRefine trades compute for robustness.
Figure~\ref{fig:refine-ablation} shows that performance improves sharply from $N=0$ (single-pass, using only the previous chunk's gate) to $N=1$, then continues to rise modestly until $N=3$, after which it saturates.
The initial jump from $N=0$ to $N=1$ reflects the value of using the current prediction to recompute the gate: the previous chunk may be stale, especially after large observation changes.
Saturation after $N=3$ suggests that the refinement loop reaches a stable self-consistent chunk quickly; later iterations have little room to change the chunk once the gate and action representation agree.
In terms of latency, each refinement iteration requires one additional forward pass through the action expert but not through the vision-language backbone, since $H_t$ and $\cR(H_t)$ are cached after the first pass.
On a single A100 GPU, one iteration adds approximately 8\,ms for $\pi_{0.5}$, so $N=3$ increases per-step inference time by roughly 24\,ms, well within the 100\,ms budget of a 10\,Hz control loop.

\noindent\textbf{Transport discrepancy stays informative.}
A potential failure mode is discrepancy collapse: if $D$ converges to a constant, the gate saturates and stops reflecting reliability.
Figure~\ref{fig:transport-curve} plots the average transport discrepancy during training.
The curve decreases as the backbone learns better representations but stabilizes at a non-trivial level rather than collapsing to zero.
This confirms that the stop-gradient on the gate prevents the model from directly minimizing $D$, preserving its diagnostic value throughout training.
Table~\ref{tab:reliability-diagnostics} connects this training-time behavior to rollout outcomes: failed rollouts have larger normalized discrepancy $\hat D$ and smaller gate values, and refining high-discrepancy states yields larger gains.

\begin{table}[tb]
\centering
\caption{Reliability diagnostics of normalized discrepancy $\hat D$ and gate $g$. Brackets report the interquartile range on Perturb-OOD.}
\label{tab:reliability-diagnostics}
\setlength{\tabcolsep}{5pt}
\resizebox{.7\linewidth}{!}{
\begin{tabular}{lcc}
\toprule
\textbf{Quantity} & \textbf{LIBERO-Long} & \textbf{Perturb-OOD} \\
\midrule
$\hat D$ success / failure
& 0.41 / 0.74
& 0.44 {\scriptsize[0.36,0.52]} / 0.79 {\scriptsize[0.69,0.88]} \\
$g$ success / failure
& 0.71 / 0.39
& 0.68 {\scriptsize[0.60,0.75]} / 0.35 {\scriptsize[0.27,0.43]} \\
AUROC$(\max_tD_t)$ / top-$D$ gain
& 0.82 / +8.1
& 0.84 / +9.3 \\
\bottomrule
\end{tabular}}
\end{table}

\begin{figure}[tb]
  \centering
  \includegraphics[width=.7\linewidth]{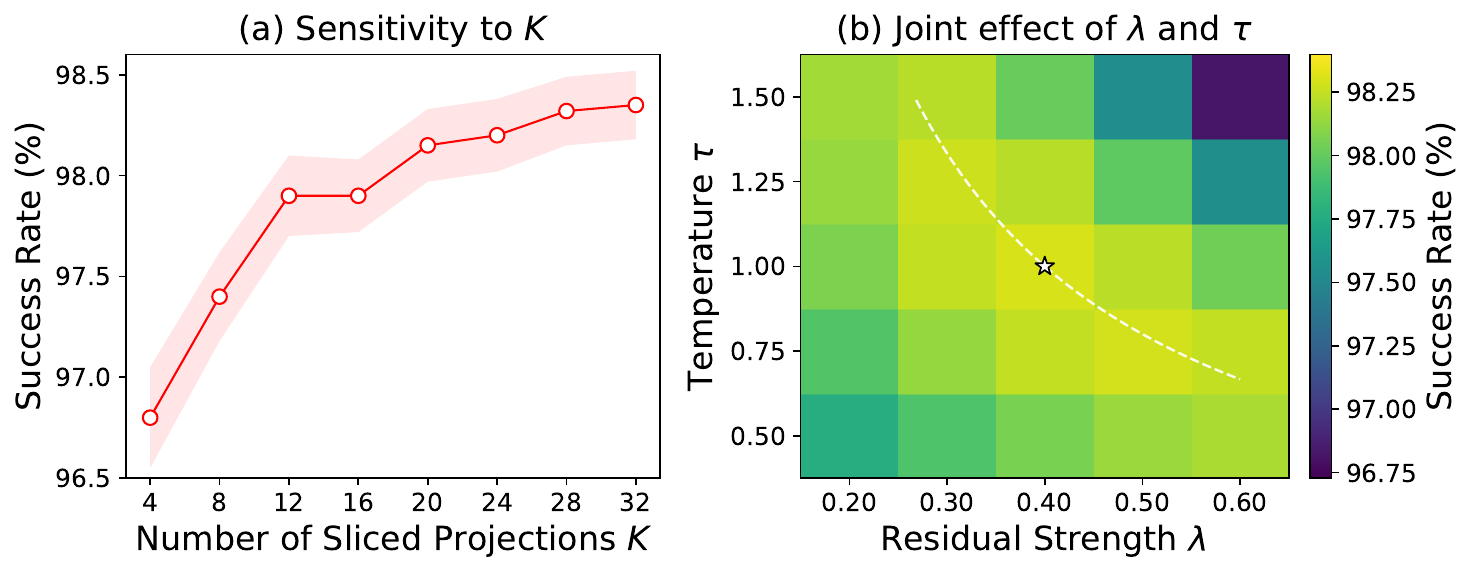}
  \caption{\textbf{Hyperparameter sensitivity.}
  Left: varying the number of sliced projections $M$.
  Right: joint sweep of refinement strength $\lambda$ and temperature $\tau$.}
  \label{fig:hyper-sens}
\end{figure}

\noindent\textbf{Hyperparameter sensitivity.}
Figure~\ref{fig:hyper-sens} reports sensitivity to the number of projections $M$ and to $(\lambda,\tau)$.
Performance is stable across a wide range of $M$: even $M=16$ projections yield most of the gain, and increasing beyond $M=32$ provides diminishing returns.
This is consistent with known convergence properties of sliced Wasserstein estimators.
The $(\lambda,\tau)$ sweep shows that moderate values ($\lambda\approx0.1$, $\tau\approx1.0$) work well.
Large $\lambda$ over-amplifies the residual correction and can destabilize training.
Small $\tau$ makes the gate nearly constant, reducing \DiG to a uniform residual correction similar to the fixed-gate ablation.
The gate clamp $g_{\min}=0.05$ prevents complete suppression of any training sample.
Setting $g_{\min}=0$ allows the gate to fully zero out high-discrepancy pairs, which risks discarding rare but valid demonstrations that happen to have unusual observation-action alignment.
Conversely, raising $g_{\min}$ above 0.2 weakens the reweighting effect and narrows the gap between coherent and shortcut samples, reducing the benefit of gating.
The default value $g_{\min}=0.05$ balances these two extremes and performs consistently across all benchmarks without per-task tuning.

\section{Conclusion}

Transport discrepancy provides a simple reliability signal for VLA policies that use flow matching in the action expert.
\DiG uses this signal to gate a lightweight residual refinement and to reweight the flow-matching loss via $\sg(\cdot)$ applied only to the gate, so the signal stays diagnostic while learning still follows the flow-matching objective.
\DiGRefine reuses the same mechanism at test time.
Results across simulation and real robots show consistent robustness gains, with the largest improvements under strong shifts and long horizons.
Extending \DiG to discrete-action models is a promising direction, as the transport discrepancy concept applies whenever observation and action representations share a common feature space.
On the practical side, \DiGRefine adds $N$ extra action-expert forward passes per step, which may matter for faster control loops.

\bibliographystyle{unsrtnat}
\bibliography{ref}

\clearpage
\appendix
\makeatletter

\renewcommand{\theHsection}{appendix.\Alph{section}}

\renewcommand{\theHsubsection}{appendix.\Alph{section}.\arabic{subsection}}
\makeatother

\section{Proofs}
\label{app:main-proofs}

This appendix provides formal proofs for Proposition~\ref{prop:posterior} and Theorem~\ref{thm:mass}.

\subsection{Proof of Proposition~\ref{prop:posterior}}

\begin{proof}
Fix a projection direction $\theta\in\R^d$ with $\|\theta\|_2=1$.
Define the projected features $u_i=\theta^\top h_i$ and the projected centroid $v=\theta^\top \bar z$.
Let $e_i=u_i-v$.

By assumption, the deviations are i.i.d.\ Gaussian:
\begin{equation}
e_i \sim \mathcal{N}(0,\sigma^2),
\qquad
i=1,\ldots,T.
\end{equation}
Normalize each deviation by $\sigma$.
Then $e_i/\sigma\sim\mathcal{N}(0,1)$.
Define the normalized sum of squares
\begin{equation}
Y \;=\; \sum_{i=1}^T \left(\frac{e_i}{\sigma}\right)^2.
\end{equation}
Since the summands are independent standard normal squares, $Y$ follows a chi-square distribution with $T$ degrees of freedom:
\begin{equation}
Y \sim \chi^2_T.
\end{equation}

Now relate $Y$ to the transport cost $D_\theta$.
By definition,
\begin{equation}
D_\theta
\;=\;
\frac{1}{T}\sum_{i=1}^T e_i^2
\;=\;
\frac{\sigma^2}{T}\,Y.
\label{eq:Dtheta-scale}
\end{equation}
The chi-square density is, for $y>0$,
\begin{equation}
p_Y(y)
\;=\;
\frac{1}{2^{T/2}\Gamma(T/2)}\,y^{T/2-1}\exp(-y/2).
\label{eq:chisq-density}
\end{equation}
Apply the change of variables implied by Eq.~(\ref{eq:Dtheta-scale}):
\begin{equation}
y \;=\; \frac{T}{\sigma^2}D_\theta,
\qquad
\frac{dy}{dD_\theta} \;=\; \frac{T}{\sigma^2}.
\end{equation}
Therefore, for $D_\theta>0$,
\begin{align}
p(D_\theta)
&=
p_Y\!\left(\frac{T}{\sigma^2}D_\theta\right)\cdot\frac{T}{\sigma^2}
\nonumber\\
&=
\frac{T}{\sigma^2}\cdot
\frac{1}{2^{T/2}\Gamma(T/2)}
\left(\frac{T}{\sigma^2}D_\theta\right)^{T/2-1}
\exp\!\left(-\frac{T}{2\sigma^2}D_\theta\right).
\label{eq:Dtheta-density}
\end{align}
Collect all factors that do not depend on $D_\theta$ into the constant
\begin{equation}
C
\;=\;
\frac{T}{\sigma^2}\cdot
\frac{1}{2^{T/2}\Gamma(T/2)}
\left(\frac{T}{\sigma^2}\right)^{T/2-1}.
\end{equation}
Then Eq.~(\ref{eq:Dtheta-density}) becomes Eq.~(\ref{eq:Dtheta-gamma}).
\end{proof}

\subsection{Proof of Theorem~\ref{thm:mass}}

\begin{proof}
Start from the mixture model in Eq.~(\ref{eq:mixture}):
\begin{equation}
P \;=\; (1-\rho)\,P_{\mathrm{coh}} + \rho\,P_{\mathrm{sh}}.
\end{equation}
By assumption, the gate is constant on each component.
It equals $g_1$ on $P_{\mathrm{coh}}$ and $g_0$ on $P_{\mathrm{sh}}$, with $g_1>g_0>0$.

Define the normalizing constant of $Q$:
\begin{equation}
Z \;=\; \E_{(H,a)\sim P}[g(H,a)].
\end{equation}
Use the mixture form of $P$ and the component-wise constancy of $g$:
\begin{align}
Z
&=
(1-\rho)\,\E_{P_{\mathrm{coh}}}[g] + \rho\,\E_{P_{\mathrm{sh}}}[g]
\nonumber\\
&=
(1-\rho)\,g_1 + \rho\,g_0.
\label{eq:Z-mass}
\end{align}

Now compute $Q$ on an arbitrary measurable set $A$.
By definition in Eq.~(\ref{eq:Qdef}),
\begin{align}
Q(A)
&=
\frac{1}{Z}\int_A g(H,a)\,P(dH\,da)
\nonumber\\
&=
\frac{1}{Z}\Big((1-\rho)\int_A g\,dP_{\mathrm{coh}} + \rho\int_A g\,dP_{\mathrm{sh}}\Big)
\nonumber\\
&=
\frac{1}{Z}\Big((1-\rho)\,g_1\,P_{\mathrm{coh}}(A) + \rho\,g_0\,P_{\mathrm{sh}}(A)\Big).
\label{eq:Q-on-A}
\end{align}
Define
\begin{equation}
\rho_Q \;=\; \frac{\rho\,g_0}{(1-\rho)\,g_1 + \rho\,g_0}.
\end{equation}
Then $1-\rho_Q=\frac{(1-\rho)\,g_1}{(1-\rho)\,g_1 + \rho\,g_0}$ and Eq.~(\ref{eq:Q-on-A}) becomes
\begin{equation}
Q(A)
=
(1-\rho_Q)\,P_{\mathrm{coh}}(A) + \rho_Q\,P_{\mathrm{sh}}(A).
\end{equation}
This proves the mixture form of $Q$ and Eq.~(\ref{eq:rhoQ}).

To show $\rho_Q<\rho$, use $g_1>g_0$:
\begin{equation}
(1-\rho)\,g_1 + \rho\,g_0
>
(1-\rho)\,g_0 + \rho\,g_0
=
g_0.
\end{equation}
Therefore,
\begin{equation}
\rho_Q
=
\frac{\rho\,g_0}{(1-\rho)\,g_1 + \rho\,g_0}
<
\frac{\rho\,g_0}{g_0}
=
\rho.
\end{equation}

Finally, prove the risk bound in Eq.~(\ref{eq:coh-bound}).
For any $\ell\in[0,1]$ and any $\theta$,
\begin{align}
\E_Q[\ell(\theta)]
&=
(1-\rho_Q)\,\E_{P_{\mathrm{coh}}}[\ell(\theta)] + \rho_Q\,\E_{P_{\mathrm{sh}}}[\ell(\theta)].
\end{align}
Subtract $\E_{P_{\mathrm{coh}}}[\ell(\theta)]$:
\begin{align}
\E_Q[\ell(\theta)] - \E_{P_{\mathrm{coh}}}[\ell(\theta)]
&=
\rho_Q\Big(\E_{P_{\mathrm{sh}}}[\ell(\theta)] - \E_{P_{\mathrm{coh}}}[\ell(\theta)]\Big).
\end{align}
Because $\ell\in[0,1]$, the difference of expectations lies in $[-1,1]$.
Taking absolute values yields
\begin{equation}
\big|\,\E_Q[\ell(\theta)] - \E_{P_{\mathrm{coh}}}[\ell(\theta)]\,\big|
\le
\rho_Q,
\end{equation}
which is Eq.~(\ref{eq:coh-bound}).
\end{proof}

\clearpage

\section{Hyperparameters}
\label{app:hyperparams}

Table~\ref{tab:hyperparams} lists the concrete values used in all experiments.
Both backbones follow their official fine-tuning recipes, and the \DiG-specific settings are shared across every benchmark and backbone.

\begin{table}[H]
\centering
\caption{Hyperparameter summary. \DiG-specific rows are identical for both backbones. Backbone rows follow the official fine-tuning recipes; $\pi_{0.5}$ uses a two-phase schedule whose per-phase batch size and step count we inherit without modification.}
\label{tab:hyperparams}
\small
\begin{tabular}{llcc}
\toprule
Symbol / Setting & Description & $\pi_{0.5}$ & GR00T-N1 \\
\midrule
$\tau$ & Gate temperature & 1.0 & 1.0 \\
$\lambda$ & Residual strength & 0.1 & 0.1 \\
$g_{\min}$ & Gate floor & 0.05 & 0.05 \\
$M$ & Sliced projections & 32 & 32 \\
$N$ & Refinement iterations & 3 & 3 \\
\midrule
$K$ & Action chunk horizon & 16 & 16 \\
$d$ & Feature dimension & 1024 & 1024 \\
Batch size & Global & \multicolumn{1}{c}{256} & 256 \\
Training steps & Total & \multicolumn{1}{c}{30k} & 30k \\
Image resolution & Input crop & $224\!\times\!224$ & $224\!\times\!224$ \\
GPU & Hardware & $8\times$A800 & $8\times$A800 \\
\bottomrule
\end{tabular}
\end{table}

\end{document}